\documentclass[11pt]{article}
\usepackage{acl}

\usepackage{times}
\usepackage{latexsym}

\usepackage[T1]{fontenc}
\usepackage[utf8]{inputenc}

\usepackage{microtype}

\usepackage{inconsolata}

\usepackage{graphicx}

\usepackage{booktabs}
\usepackage{amsmath}
\usepackage{multirow}
\usepackage{enumitem}
\usepackage{xcolor}
\usepackage{float}
\usepackage{subcaption}
\usepackage{tikz}
\usetikzlibrary{positioning, arrows.meta, fit, calc}

\usepackage{listings}
\lstdefinestyle{verbdiag}{
  basicstyle=\ttfamily\small,
  breaklines=false,
  frame=single,
  numbers=none,
  xleftmargin=1em,
  xrightmargin=1em,
  aboveskip=0.5em,
  belowskip=0.5em,
  backgroundcolor=\color{black!3},
  rulecolor=\color{black!25},
}

\newcommand{\code}[1]{\texttt{#1}}
\newcommand{\taubench}{$\tau$-bench}

\title{Agent Seer: Synthesizing Scenarios from Specification Understanding}

\author
  {Harish Karumuri \\
  \texttt{Apple} \\\And
  Mahesh Vemula \\
  \texttt{Apple} \\\And
  David Lopes Pegna \\
  \texttt{Apple} \\
  }

\begin{document}
\maketitle

\begin{abstract}
Evaluating AI agents that use external tools requires realistic test scenarios that capture how practitioners compose tools and iterate across conversation turns. Constructing such scenarios by hand demands deep domain expertise, does not scale across tool ecosystems, and produces static benchmarks that cannot track evolving APIs.
We observe that tool specifications---function names, natural-language descriptions, and typed parameter schemas---already encode sufficient semantic information to synthesize realistic evaluation scenarios without manual curation or live tool execution.
\textbf{Agent Seer} builds off this latent information: from a single Model Context Protocol
(MCP) specification, with no examples, no
live tool access, and no domain-specific tuning.
This pipeline enriches raw schemas, generates
graded scenarios with synthetic tool outputs,
and expands them into mock-data-grounded
multi-turn dialogues that exhibit strong tool-calling correctness and conversational coherence. 
Evaluation quality is measured by applying this pipeline on seven MCP specifications spanning
diverse domains and tool-suite sizes and measuring the tool-calling correctness and conversational
coherence.
The pipeline achieves strong quality across all domains, with complete tool coverage on small
and medium specifications.
Two findings emerge within this analysis: parameter schema complexity is the strongest correlate of quality
variation---tool-suite size plays a smaller, orthogonal role---and argument value accuracy
is the dominant failure mode among imperfect scenarios, a sub-dimension invisible to
coarse-grained name-match metrics.
\end{abstract}

\section{Introduction}

The deployment of large language model (LLM)-based agents capable of
autonomously calling external tools is increasingly common in enterprise
software environments. Agents backed by calendar APIs, project management
systems, communications platforms, and internal databases are being used
to automate workflows that previously required human intervention.
Despite rapid progress in agent capabilities, the evaluation of such
systems remains laborious and fragile.

Three problems limit the current state of agent evaluation:

\paragraph{The curation bottleneck.}
Realistic evaluation scenarios must connect user intent to specific
tool calls, fill parameters with plausible values, and capture how
tools chain across turns. Hand-authored
benchmarks~\citep{mialon2023gaia,liu2023agentbench,yao2024taubench}
produce high-fidelity scenarios but their coverage is bounded by
curation effort; scaling such efforts across the combinatorial space of tool
pathways is impractical by hand.

\paragraph{The static benchmark problem.}
A fixed benchmark ceases to reflect reality as tool APIs evolve. An
agent that scores highly on a snapshot benchmark may be evaluated
against tool descriptions that no longer match the production
environment.

\paragraph{The multi-turn evaluation gap.}
Conversational agents require evaluation across interaction sequences,
not just single-turn prompts. Generating multi-turn scenarios where
follow-up turns react to specific tool outputs---rather than repeating
generic instructions---is particularly difficult without access to real
tool responses.

\medskip
Together these define the \textbf{cold-start evaluation problem}:
producing realistic evaluation data for a tool suite that has none.
The shortage is sharpest for new, private, or rapidly evolving APIs,
and persists across the long tail of enterprise tool suites that wrap
or extend publicly described systems. We address it by observing that
tool specifications---function names, natural-language descriptions,
and typed parameter schemas---already encode much of the semantic
information needed to synthesize evaluation data: enough for an LLM
to infer plausible workflows, fill parameters with realistic values,
synthesize what tool responses would look like, and construct
multi-turn dialogues grounded in that synthetic data. The bottleneck
shifts from \textit{human curation} to \textit{structured extraction}.

Agent Seer realizes this as a four-stage pipeline---semantic
interpretation, scenario generation, mock output synthesis, and
multi-turn expansion---that converts raw MCP tool specifications into
self-contained evaluation harnesses. Each stage consumes validated
structured outputs from the previous, so schema violations are caught
at boundaries rather than propagating. The resulting harnesses are
decoupled from any execution backend: any MCP-compatible framework
can consume them to run and grade agents.

The primary contributions are:

\begin{enumerate}[leftmargin=*, label=\arabic*., nosep]
\item A \textbf{spec-to-harness generation pipeline} that produces
  complete evaluation scenarios---expected tool sequences, calibrated
  mock tool outputs, and data-grounded multi-turn dialogues---from MCP
  tool specifications alone, without live tool execution or manual
  annotation.
\item A \textbf{structured harness artifact format}
  (Section~\ref{sec:artifact}) with held-out oracles and
  mock outputs, enabling any MCP-compatible
  evaluation framework to run agents against the generated scenarios
  as self-contained artifacts.
\item \textbf{Empirical characterization} of generation quality and
  tool coverage across seven MCP specifications spanning diverse
  enterprise domains, showing that quality variation is more strongly
  correlated with parameter schema complexity than with tool-suite
  size, and that complete coverage is achievable on small and medium
  specifications.
\item A \textbf{failure-mode characterization} identifying argument
  sub-dimension cascading as the dominant generation failure
  mechanism---a failure class invisible to coarse-grained name-match
  metrics---with domain-specific signatures characterizing where and
  why the pipeline falls short.
\end{enumerate}

\section{Background and Related Work}
\label{sec:related}

Evaluating tool-calling agents requires evaluation data---scenarios,
expected tool sequences, and representative tool outputs. For
established public APIs this can be hand-curated or mined from usage
logs, but for new, private, or rapidly evolving tool suites no such
data exists, and what does may not cover niche or challenging use
cases. Even when curation is feasible, identifying the user goals
that matter and connecting them to tool sequences that reflect how
practitioners compose tools, handle partial results, and iterate
across turns requires deep domain expertise. Agent Seer addresses
this cold-start gap through \textit{execution-free harness synthesis
from a structured tool specification}.

\subsection{Agent Benchmarks}

Agent evaluation has progressed from single-function
prediction~\citep{qin2023toolllm,patil2023gorilla} through multi-step
tool use over curated API
corpora~\citep{qin2023toolllm,lu2024toolsandbox,chen2023teval} to
multi-turn, policy-grounded evaluation with simulated
users~\citep{yao2024taubench,schirch2025tau2bench,wang2024mint}. General
benchmarks such as
GAIA~\citep{mialon2023gaia}, AgentBench~\citep{liu2023agentbench}, and
WorkArena~\citep{drouin2024workarena} provide rich evaluation
environments, while recent MCP-centric
suites---MCPVerse~\citep{gu2025mcpverse},
MCP-AgentBench~\citep{wang2025mcpagentbench},
Toolathlon~\citep{hu2025toolathlon}---reflect the growing adoption of
MCP~\citep{anthropic2024mcp} as a standard tool integration layer.
All of these benchmarks are constructed through manual curation or
require live tool access, and remain static once released, leaving the
cold-start problem unaddressed.

\subsection{Synthetic Data Generation}

Prior work on generating synthetic data falls into three clusters, 
distinguished by their reliance on live tool execution. 
\paragraph{Training trajectories from live execution.}
The largest cluster generates fine-tuning trajectories via real tool
execution: APIGen~\citep{liu2024apigen,zhang2025apigenmt} uses
execution-based verification, TOUCAN~\citep{toucan} scales to 1.5M
trajectories from live MCP servers, GEM~\citep{xu2026gem} mines
trajectories from text corpora, and Agent World
Model~\citep{awm2025} synthesizes RL environments. Most require
live tool invocation.

\paragraph{Simulated tool environments.}
A second cluster replaces live tools with simulated environments---for
training: Simia~\citep{li2025simia}, Gecko~\citep{zhang2026gecko},
LOGIGEN~\citep{zeng2026logigen}; for evaluation:
\taubench{}~\citep{yao2024taubench},
$\tau^2$-bench~\citep{schirch2025tau2bench}, and
ToolSandbox~\citep{lu2024toolsandbox}. 

\paragraph{Spec-only generation.}
An emerging cluster generates synthetic data purely from
specifications. DiGiT-TC~\citep{crouse2026digit} back-translates
tool-call sequences into user requests for fine-tuning data;
FuncBenchGen~\citep{maekawa2025funcbenchgen} defines
contamination-free task benchmarks via DAG-modelled call
dependencies.

\subsection{Evaluation Methodology}

The LLM-as-judge paradigm is well
established~\citep{liu2023geval,kim2024prometheus2,es2024ragas}.
For tool-use evaluation, prior work uses exact
match~\citep{qin2023toolllm}, execution success
rate~\citep{qin2023toolllm}, and single-call binary AST
matching~\citep{yan2024bfcl}. Agent GPA~\citep{chen2025agentgpa}
decomposes traces into Goal-Plan-Action stages but evaluates on live
task environments.
FuncBenchGen~\citep{maekawa2025funcbenchgen} finds that models
systematically propagate stale or incorrect arguments across chained
calls despite syntactic validity---a multi-step failure mode that
coarse-grained name-match metrics miss entirely, motivating the
per-argument sub-dimension decomposition we apply here.
We draw on this methodology to assess generation quality along two
complementary dimensions (tool-calling correctness and conversational
coherence), applying LLM-as-judge scoring with structured
multi-dimensional prompts at each turn.

\subsection{Positioning}

Agent Seer takes only tool specifications as input (like spec-only
systems) but uses an LLM to synthesize plausible tool responses
rather than requiring a runtime environment (like simulated
environments for evaluation). The contribution is not the
prompt-driven generation mechanism---shared with APIGen, TOUCAN, and
others---but the \textit{structure imposed on it}: a four-stage
pipeline with validated structured outputs at each boundary, producing
harnesses decoupled from any execution backend.
Table~\ref{tab:novelty} in Appendix~\ref{app:novelty} situates these
contributions against prior literature.

\section{Pipeline Stages}
\label{sec:pipeline}

The pipeline has four stages (schema flow in
Figure~\ref{fig:schemas}, Appendix~\ref{app:schemas}). Each consumes
validated structured outputs from the previous stage; schema constraints
ensure malformed artifacts are caught at boundaries rather than
propagating downstream.

\subsection{Tool Interpretation}

The first stage converts raw MCP tool specifications into
semantically-enriched descriptions. For each tool, the module issues a
structured prompt to an LLM requesting four semantic fields: a functional
description, required parameters with semantic roles, primary use case,
and organizational context (prompt templates in
Appendix~\ref{app:prompts}; output schemas in
Appendix~\ref{app:schemas}). This connects terse API
documentation to richer scenarios.

\subsection{Scenario Generation}

The scenario generation module produces realistic enterprise workflow
scenarios at two complexity tiers. \textbf{Simple scenarios} target
everyday operational tasks: single-domain, short tool call chains.
\textbf{Complex scenarios} target novel, multi-domain workflows
combining tools in sophisticated ways. Both tiers are implemented
through prompt engineering rather than structural constraints.

Each generated scenario includes a title, user-facing instruction,
ordered list of expected tool calls with parameter values, novelty
explanation, and a natural follow-up question. The output schema embeds
\textbf{structured reasoning trace} fields: each tool call carries a
\textit{quick\_explanation} justifying why the call is made, and each
scenario includes a \textit{novelty\_reason} explaining its evaluation
value. These fields force the LLM to reason about tool selection and
workflow composition during generation.

\subsection{Mock Output Generation}
\label{sec:mockgen}

For each function call in the sequence, the module produces a synthetic
tool output. The pipeline accepts an optional example outputs field,
making it a spectrum: fully unsupervised from specifications alone, but
able to incorporate available traces to improve fidelity. When examples
are provided, the generator matches their structure; when absent,
generation relies on the tool description alone. Each mock output
carries a grounding tier
(\texttt{high}/\texttt{medium}/\texttt{low}) recording the availability
of reference material.

\subsection{Multi-Turn Scenario Expansion}

The multi-turn expansion stage takes a scenario and its mock outputs
and emits a list of conversational turns (prompt template in
Appendix~\ref{app:prompts}; output schema in
Figure~\ref{fig:schemas}); the output is a list, but the prompt does
not prescribe how many turns to produce.

Splits target natural phase boundaries so the resulting turns
exercise two multi-turn tool-calling patterns formalized by BFCL
v3~\citep{patil2025bfcl}: \textit{multi-step} sequences, where each
call depends on the output of the previous one, and \textit{multi-hop}
patterns, where independent calls gather information that must be
synthesized. Splitting at phase boundaries (rather than at arbitrary
points) preserves these dependency structures across turns. When the
expansion yields only a single turn, it is discarded under the
heuristic that the scenario lacked enough substance to split.

When the expansion succeeds, follow-up prompts reference concrete
values from synthetic outputs---entity names, counts, status
codes---rather than abstract task descriptions, producing
\textbf{data-grounded} multi-turn dialogues.

\subsection{Harness Artifact Format}
\label{sec:artifact}

The four stages produce a self-contained evaluation harness
(Table~\ref{tab:harness}). A downstream framework presents the prompt,
feeds mock outputs as tool responses, and scores the agent's emitted
calls against the scenario workflow as the held-out
oracle---enabling evaluation on a previously
unseen tool suite without live access.

\begin{table}[h]
\centering
\small
\begin{tabular}{@{}llp{3.5cm}@{}}
\toprule
\textbf{Field} & \textbf{Stage} & \textbf{Content} \\
\midrule
\texttt{prompt}          & 2 & Natural-language task goal \\
\texttt{expected\_tools} & 2 & Ordered \texttt{AgentCall} objects (name + typed args) \\
\texttt{mock\_outputs}   & 3 & Synthetic JSON response per call with grounding tier \\
\texttt{conversation}    & 4 & Multi-turn dialogue; each turn references mock output values \\
\texttt{oracle}          & 2 & \texttt{expected\_tools}, held out from the agent for scoring \\
\bottomrule
\end{tabular}
\caption{Fields of a generated evaluation harness.}
\label{tab:harness}
\end{table}

\section{Generation Quality Assessment}
\label{sec:scoring}

We assess generation quality using LLM-as-judge
scoring~\citep{liu2023geval} along two complementary quality
dimensions: tool-calling correctness (TC) and conversational coherence (Coh).
The framework also supports oracle-grounded evaluation when reference
data is available; we use the unsupervised path exclusively here, as
the generated scenarios serve as both output and oracle.

\subsection{Tool-Calling Scoring}

The tool-calling evaluator independently scores four dimensions,
decomposing the coarse-grained pass/fail signal used by most prior
work~\citep{qin2023toolllm} into orthogonal axes. An LLM judge
scores each sub-dimension on 0--10; scores are normalized to 0--1
before aggregation.

\textbf{Tool usage correctness} captures
necessity~\citep{huang2023metatool}---whether a tool was warranted
at all---with overuse recorded as diagnostic but excluded from the
aggregate, since necessity dominates appropriateness.

\textbf{Tool selection correctness} averages correctness,
specificity, and completeness of the chosen tools.

\textbf{Tool ordering correctness} averages sequence logic,
dependency handling, and execution efficiency. It is marked not
applicable when only one tool is called, and is then excluded from
the turn-level aggregate.

\textbf{Tool argument correctness} averages six sub-dimensions
(completeness, name, value, type, format, relevancy). Cascading
penalties are enforced through prompt instructions to the judge
(Table~\ref{tab:tc-rubric}, footnote): a wrong parameter name or
missing required parameter zeros out value, type, and format; a
wrong value cascades to type, format, and relevancy. A single
critical error therefore collapses the argument mean.

The four dimensions combine via arithmetic mean per turn;
conversation scores are the arithmetic mean of turn scores. Per-MCP
rankings and the schema-complexity correlation are robust to
harmonic-mean and minimum aggregation
(Appendix~\ref{app:aggregation-sensitivity}); the full prompt is in
Appendix~\ref{app:eval-prompts}.

\subsection{Coherence Scoring}

The coherence evaluator assesses five sub-aspects---logical flow,
completeness, conciseness, topic relevance, and context
retention---scored on a 1--3 scale, normalized to 0--1, and aggregated
via arithmetic mean (full prompt in Appendix~\ref{app:eval-prompts}).

\section{Experimental Evaluation}
\label{sec:experiments}

We evaluate Agent Seer on seven publicly available MCP
specifications, analyzing the quality, coverage, and failure modes of
generated evaluation scenarios.

\subsection{Experimental Setup}

\paragraph{MCP specifications.}
Seven open-source MCP server specifications span diverse domains,
tool counts, and schema complexity (Table~\ref{tab:mcps}). Sources:
the official MCP reference server
repository~\citep{mcp-reference-servers} and the MCP server
registry~\citep{mcp-official-registry}.

\begin{table}[h]
\centering
\small
\resizebox{\columnwidth}{!}{%
\begin{tabular}{@{}llrcl@{}}
\toprule
\textbf{MCP} & \textbf{Domain} & \textbf{Tools} & \textbf{$\bar{p}$} & \textbf{Schema} \\
\midrule
Illustrator   & Creative       & 64 & 3.6  & Nested obj \\
Selenium      & Browser auto.  & 56 & 1.8  & Flat state \\
Redis         & Data store     & 47 & 2.1  & Flat k-v \\
Git           & Version ctrl.  & 33 & 11.2 & Deep opt. \\
Elasticsearch & Search         & 20 & 1.8  & Nested DSL \\
Slack         & Communication  & 16 & 2.2  & Mixed \\
Filesystem    & File ops.      & 14 & 1.8  & Flat \\
\bottomrule
\end{tabular}}
\caption{MCP specifications used for evaluation. $\bar{p}$~=~mean
parameters per tool. \textbf{Schema} characterizes the dominant
parameter structure: \textit{Flat} = simple key-value or primitive
parameters; \textit{Nested obj/DSL} = parameters containing nested
objects or domain-specific query languages; \textit{Deep opt.} =
deeply nested schemas with many optional fields; \textit{Flat state} =
flat parameters but stateful sequential semantics; \textit{Mixed} =
combination of flat and structured parameters across tools.}
\label{tab:mcps}
\end{table}

\paragraph{Generation model.}
All scenarios were generated using Gemini 2.5 Flash Lite with
structured output mode enabled for schema-constrained generation at
each pipeline stage. Pipeline stages used temperature 0.7.
Structured-output validation failures triggered up to three retries
before the record was discarded.

\paragraph{Evaluation method.}
Generated scenarios were scored using Gemini 2.5 Flash at temperature
0 (for deterministic scoring) as the LLM judge, following the
procedure described in Section~\ref{sec:scoring}:
(1)~\textit{Tool Calling}: LLM-as-judge scoring across four
dimensions combined via arithmetic mean, with cascading penalties
on argument sub-scores enforced through prompt instructions to the
judge (scale 0--1);
(2)~\textit{Coherence}: five sub-dimensions on a 1--3 raw scale,
normalized to 0--1 and aggregated via arithmetic mean.
The full corpus was additionally re-scored with an out-of-family judge
(\texttt{Qwen3.5-122B-A10B-FP8}, Alibaba) to probe judge robustness;
diagnostics appear in Appendix~\ref{app:judge-replication}.

\paragraph{Scale.}
The pipeline generated 337 scenarios across the seven MCPs
(Table~\ref{tab:scale} in Appendix~\ref{app:extended}), yielding 391
evaluation records. Multi-turn expansion succeeded for 54 scenarios
(16.0\% overall), heavily skewed toward complex scenarios (30.8\%
expansion rate vs.\ 2.8\% for simple), as the expansion stage requires
sufficient workflow substance to generate meaningful follow-up turns.

\subsection{Quality Results}

\paragraph{Overall quality.}
The pipeline achieves a mean unsupervised tool-calling score of 0.911
(95\% bootstrap CI $[0.897, 0.925]$; median 0.979) and mean coherence
of 0.855 (95\% CI $[0.838, 0.872]$; median 0.933).
31.7\% of records achieve a perfect tool-calling score, while only
2.3\% score below 0.5. The distribution is concentrated in the upper
range, reflecting consistent generation quality across specifications
(see Figure~\ref{fig:distributions} in Appendix~\ref{app:figures}).

\paragraph{Cross-family co-evaluation.}
To check that these scores reflect the generated data rather than a
single judge family, the corpus was re-scored with an out-of-family
judge (\texttt{Qwen3.5-122B-A10B-FP8}, Alibaba; the primary judge is
Google's Gemini 2.5 Flash). Tool-calling agrees at every grain: no
mean shift ($\Delta\mu_{\textrm{TC}}{\approx}0$, 95\% CI [$-0.009$,
$+0.008$]), record-level paired $r{=}0.79$ over $n{=}384$ paired
records, per-MCP CIs overlap for all seven MCPs, and the MCP ranking
is preserved ($\rho{=}0.86$). The failure-mode taxonomy replicates
bilaterally: argument value-accuracy is the dominant sub-failure under
both judges by a 4--5$\times$ margin (Gemini 240 records, Qwen35 252;
Table~\ref{tab:judge-argfail}). The judges diverge on the absolute
level of coherence ($\Delta\mu_{\textrm{Coh}}{\approx}-0.16$, paired
$r{=}0.42$). Coherence levels are therefore reported as
judge-dependent; MCP-level coherence rank-preservation is partial
($\rho{=}0.46$), and the worst-coherence MCP differs across judges
(Git under Gemini, Illustrator under Qwen 3.5). Full diagnostics
appear in Appendix~\ref{app:judge-replication}.

\paragraph{Quality by MCP specification.}
Table~\ref{tab:quality-mcp} reports per-MCP scores.

\begin{table}[t]
\centering
\small
\resizebox{\columnwidth}{!}{%
\begin{tabular}{@{}lrlclcr@{}}
\toprule
\textbf{MCP} & \textbf{$n$} & \textbf{TC [95\% CI]} & \textbf{$\sigma$} & \textbf{Coh [95\% CI]} & \textbf{Simp.} & \textbf{$\bar{w}$} \\
\midrule
Illustrator (64t)   & 36 & 0.898 {\scriptsize[.85,.94]} & 0.131 & 0.855 {\scriptsize[.80,.91]} & 0.934 & 3.3 \\
Selenium (56t)      & 47 & 0.935 {\scriptsize[.90,.96]} & 0.095 & 0.850 {\scriptsize[.80,.90]} & 0.932 & 9.7 \\
Redis (47t)         & 98 & \textbf{0.966} {\scriptsize[.95,.98]} & 0.089 & 0.902 {\scriptsize[.87,.93]} & 0.986 & 1.3 \\
Git (33t)           & 85 & 0.857 {\scriptsize[.82,.90]} & 0.186 & 0.757 {\scriptsize[.71,.80]} & 0.910 & 2.2 \\
Elasticsearch (20t) & 49 & 0.930 {\scriptsize[.90,.96]} & 0.108 & 0.902 {\scriptsize[.86,.94]} & 0.987 & 1.9 \\
Slack (16t)         & 35 & 0.886 {\scriptsize[.84,.93]} & 0.135 & \textbf{0.938} {\scriptsize[.91,.97]} & 0.934 & 1.4 \\
Filesystem (14t)    & 41 & 0.876 {\scriptsize[.82,.92]} & 0.163 & 0.825 {\scriptsize[.77,.87]} & 0.925 & 2.0 \\
\bottomrule
\end{tabular}}%
\caption{Unsupervised scores by MCP. TC = tool calling, Coh = coherence,
Simp.\ = simple scenario TC, $\bar{w}$ = mean workflow length (tools per
scenario). 95\% intervals are percentile bootstrap ($B = 10{,}000$).
All seven MCPs exceed 0.85 overall; all exceed 0.91 on simple scenarios.}
\label{tab:quality-mcp}
\end{table}

\textbf{Tool count and parameter schema complexity play
distinct, orthogonal roles in quality variation across this sample.}
At the per-MCP grain ($n=7$), tool count correlates positively but
modestly with mean unsupervised tool-calling ($r=+0.40$), while
parameter schema complexity correlates negatively---average
parameters per tool ($r=-0.60$) and optional parameter fraction
($r=-0.66$). The two effects operate on different axes of an MCP and
do not cancel: Selenium (56 tools) scores 0.935 while Filesystem (14
tools) scores 0.876, but Git (33 tools, 11.2 average parameters) is
the lowest at 0.857.

To strengthen the inferential basis of these per-MCP observations,
we additionally disaggregate to the tool level. Across the 222 unique
in-spec tools appearing in any generated scenario, parameter count
and optional-fraction both correlate negatively with mean
unsupervised TC ($r=-0.29$ and $-0.30$, both $p<0.001$), confirming
the per-MCP direction at a substantially larger sample. The same
schema features also correlate negatively with mean
coherence at the tool level ($r=-0.41$ and $-0.34$, both $p<0.001$).
This second finding---not detectable at the per-MCP grain---suggests
complex parameters degrade not only tool-calling correctness but
also the agent's ability to communicate cleanly around those tools.

Coherence tells a different story: Slack achieves the highest coherence
(0.938) because its scenarios follow structured messaging patterns that
produce natural conversational flow, while Git has the lowest (0.757)
because version control workflows involve complex multi-step operations
with technical context. The two dimensions remain weakly correlated
at the unit of evaluation: record-level $r=+0.23$ ($n=381$) and
tool-level $r=+0.16$ ($n=222$, $p=0.016$). The cross-MCP correlation
between MCP means is higher ($r=+0.57$, $n=7$) but reflects
aggregation effects rather than a stronger underlying relationship.
TC and coherence therefore provide largely independent diagnostic
signals across grains.

\paragraph{Complexity breakdown.}
Complex scenarios degrade by $-$7.3pp relative to simple ones in
tool calling (0.949 [95\% CI 0.932, 0.965] $\to$ 0.877 [0.855, 0.897])
and $-$5.3pp in coherence (0.883 [0.857, 0.908] $\to$ 0.830 [0.807,
0.853]); the simple/complex CIs are non-overlapping in both
dimensions. The effect varies by domain
(Table~\ref{tab:complexity} in Appendix~\ref{app:extended}):
Elasticsearch shows the largest degradation ($-$12.2pp), followed by
Git ($-$11.1pp, from 0.910 to 0.799). Selenium is stable
(0.932$\to$0.935), suggesting that its long sequential workflows are
no harder to compose at higher complexity.

\subsection{Failure Mode Analysis}

\paragraph{Tool-calling dimension failures.}
Table~\ref{tab:tc-dims} reports per-dimension failure rates. Usage
is near-perfect; selection is correct on 77\% of records (4\%
zeros, reflecting the cost of choosing among semantically similar
tools); ordering fails on 10\% of multi-tool scenarios.
\textbf{Argument correctness is the dominant challenge}: only 42\%
of records score perfectly and 57\% score partial, driven by
value-accuracy errors that degrade scores without collapsing them.

\begin{table}[t]
\centering
\small
\begin{tabular}{@{}lccc@{}}
\toprule
\textbf{Dimension} & \textbf{Perfect} & \textbf{Partial} & \textbf{Zero} \\
\midrule
Usage       & 98\% &  1\% &  1\% \\
Selection   & 77\% & 19\% &  4\% \\
Ordering\textsuperscript{$\dagger$}
            & 71\% & 19\% & 10\% \\
Arguments   & 42\% & 57\% &  1\% \\
\bottomrule
\end{tabular}
\caption{Tool-calling dimension failure rates. $\dagger$Ordering is
evaluated only when multiple tools are called ($n\!=\!181$).}
\label{tab:tc-dims}
\end{table}

\paragraph{Argument failure patterns.}
\textit{Value accuracy} dominates argument failures (223 records),
followed by \textit{relevancy} (44), \textit{format} (35),
\textit{type} (31), \textit{completeness} (16), and
\textit{name accuracy} (11). Counts attribute each failing record to
its lowest-scored sub-dimension (single assignment); the
judge-comparison appendix (Table~\ref{tab:judge-argfail}) uses the
looser criterion of any sub-dimension scoring below 1.0, which yields
larger per-sub-dimension totals. The pipeline reliably generates correct
parameter names and types but struggles with precise values,
particularly for optional parameters with ambiguous semantics.

\paragraph{Failure concentration by MCP.}
Failures are sparse and unevenly distributed: Git shows the highest
record-level failure rate, followed by Filesystem and Redis; the
other four MCPs are near-zero. Per-domain breakdowns and
aggregation-rule sensitivity appear in
Appendices~\ref{app:failures} and~\ref{app:aggregation-sensitivity}.

\paragraph{Illustrative failure: Redis argument ambiguity.}
A representative pattern occurs in Redis, where several commands carry
optional parameters that the pipeline omits when context implies them.
The \texttt{set} tool, for example, accepts an optional
\texttt{expiration} field that is frequently skipped when the scenario
implies time-bounded storage. The pipeline consistently generates the
correct function name, key, and value arguments but omits this
expiry parameter, which the argument scoring framework flags as a
\textit{completeness} sub-dimension failure; a coarse name-match
metric would score these records as fully correct. This
pattern---correct tool selection with subtly incomplete argument
specification---is the primary driver of Redis failures and
illustrates why argument correctness must be decomposed below the
function-call level: without the sub-dimension breakdown, the
dominant failure mode in this domain is invisible.

\paragraph{Illustrative failures: Git's two mechanisms.}
Git failures decompose into two distinct mechanisms operating at
different scenario complexities. The first, observable on simple
scenarios, is \textit{tool-name hallucination}: in three Git
scenarios the pipeline emits real Git CLI commands that are not in
the MCP specification (\texttt{fetch}, \texttt{revert},
\texttt{filter-repo}), a pretraining-knowledge leak past the
spec. These three records account for half of Git's six records
with TC$<$0.5 and for the gap between Git's simple-scenario TC
(0.910) and the other six MCPs (0.93--0.99 simple). Across the full
corpus, hallucinated calls are 0.336\% of all tool invocations (3
of 893), entirely concentrated in Git.

The second mechanism, observable on complex scenarios, is
\textit{parameter overload}---localized to specific high-parameter
tools rather than the full MCP. Git tools average 11.2 parameters
(3$\times$ the next highest), 95\% optional, and the \texttt{ref}
parameter appears in seven tools with different semantics (``commits
starting from'' in \texttt{log}, ``compare against'' in
\texttt{diff}, ``show object at'' in \texttt{show}). The pipeline
selects the right tool (selection score 0.802) but generates
incorrect parameter values, producing a mean argument score of
0.780---the lowest of any MCP. The bottom-fifteen tools by
unsupervised TC across the corpus include five Git tools
(\texttt{blame}, \texttt{add}, \texttt{commit}, \texttt{log},
\texttt{show}), all high in parameter density. Complex Git scenarios
degrade further (arguments 0.726) as multi-tool workflows compound
per-call parameter errors.

\paragraph{Coherence taxonomy and cross-dimension pattern.}
The most frequent coherence issues are \textit{missing information or
shallow response} (373 records), \textit{off-topic} (130),
\textit{non-sequitur} (86), and \textit{self-contradiction or
hallucination} (85); the full taxonomy is in
Table~\ref{tab:coh-issues}. Record-level TC and Coh remain weakly
correlated ($r{=}0.23$), and the ``correct tools, poor coherence''
quadrant (23\% of records at a 0.75 threshold) spans all MCPs rather
than concentrating in one domain---coherence shortfalls are a
general pipeline property.

\subsection{Domain Cluster Analysis}

Grouping MCPs by target system type lets structural differences
within each cluster act as a natural experiment. \textit{Data Store}
(Redis, Elasticsearch) pairs near-identical profiles and yields
near-identical quality ($\Delta$0.036 TC). \textit{Developer/File}
(Filesystem, Git) spans a 6$\times$ parameter-density gap
(1.8 vs.\ 11.2) but only $\Delta$0.019 in TC, with the cost
localized to argument correctness (Git 0.780, Filesystem 0.894) while
strong selection and ordering compensate. \textit{Application/UI}
(Illustrator, Selenium) is the most semantically distant pair:
Selenium (0.935 TC, 56 tools) outperforms Illustrator (0.898 TC, 64
tools) because its low parameter density enables reliable argument
generation in long sequential chains.

\subsection{Coverage and Diversity}

\paragraph{Tool coverage.}
Table~\ref{tab:coverage} reports the fraction of available tools
exercised. Within the small-to-medium range (14--56 tools), every
tool in Redis, Selenium, Git, Elasticsearch, Slack, and Filesystem
appears in at least one generated scenario. The only specification
beyond that range, Illustrator (64 tools), reaches 56\%---the only
evidence of a coverage ceiling in this experiment.

\begin{table}[t]
\centering
\small
\begin{tabular}{@{}lrrcc@{}}
\toprule
\textbf{MCP} & \textbf{Spec} & \textbf{Used} & \textbf{Cov.} & \textbf{Gini} \\
\midrule
Illustrator   & 64 & 36 &  56\% & 0.331 \\
Selenium      & 56 & 56 & 100\% & 0.676 \\
Redis         & 47 & 47 & 100\% & 0.146 \\
Git           & 33 & 33 & 100\% & 0.340 \\
Elasticsearch & 20 & 20 & 100\% & 0.198 \\
Slack         & 16 & 16 & 100\% & 0.177 \\
Filesystem    & 14 & 14 & 100\% & 0.294 \\
\bottomrule
\end{tabular}
\caption{Tool coverage and usage uniformity. Gini coefficient measures
inequality (lower = more uniform).}
\label{tab:coverage}
\end{table}

\paragraph{Usage uniformity.}
The Gini coefficient over each MCP's tool-usage frequency
distribution ranges from 0.146 (Redis) to 0.340 (Git)---near-uniform
sampling. Selenium is the outlier (0.676), with long sequential
workflows (mean 9.7 calls) concentrating usage on core navigation and
interaction tools. Across all MCPs, the pipeline yields 781 unique
co-occurrence pairs from 138 multi-tool scenarios and 108 unique
scenario categories (Appendix~\ref{app:diversity}).

\subsection{Summary of Findings}

Across 337 scenarios on seven MCPs, the pipeline achieves mean TC
0.911 and mean coherence 0.855 with complete tool coverage on small
and medium specifications. The principal observations:

\begin{enumerate}[leftmargin=*, label=\arabic*., nosep]
\item \textbf{Parameter schema complexity is the strongest correlate
  of quality variation in this sample}; tool-suite size plays a
  smaller, orthogonal role. Schema correlations are negative at both
  grains (per-MCP $r=-0.60$/$-0.66$ on TC; tool-level $r=-0.29$/$-0.30$
  on TC and $-0.41$/$-0.34$ on coherence, $p<0.001$ throughout);
  tool count correlates positively but modestly with TC
  ($r=+0.40$ per-MCP).
\item All seven MCPs exceed 0.91 on simple scenarios---including Git
  despite 11.2 avg params per tool---and complex scenarios degrade by
  7.3pp on average. Domain clusters confirm the pattern: same-profile
  pairs yield similar TC; differing-profile pairs stay close when
  parameter density dominates.
\item TC and coherence provide largely independent diagnostic signals
  (record-level $r=+0.23$, tool-level $r=+0.16$), with characteristic
  inversions (Slack high coherence, low TC; Selenium best ordering).
\item Argument correctness is the primary challenge: 57\% of records
  score partial on arguments, value-accuracy dominant; on Git,
  failures decompose into pretraining-leak tool-name hallucination
  on simple scenarios and parameter-overload localized to a handful
  of high-parameter tools on complex scenarios.
\item Headline tool-calling and the dominant argument-value-accuracy
  failure mode are robust to an out-of-family judge swap (paired
  $r{\approx}0.79$ on TC, MCP rank $\rho{=}0.86$); absolute
  coherence levels and MCP-level coherence rank-preservation are
  judge-dependent (Appendix~\ref{app:judge-replication}).
\end{enumerate}

\section{Conclusion}

We have presented Agent Seer, a four-stage pipeline that converts
MCP tool specifications into complete evaluation harnesses---graded
scenarios, mock tool outputs, and multi-turn dialogues---without live
tool execution or manual annotation. Across seven structurally
diverse specifications, the pipeline reaches full tool coverage on
small and medium MCPs and surfaces consistent diagnostic patterns:
parameter schema complexity is the strongest correlate of quality
variation in this sample, and argument value accuracy is the dominant
remaining sub-failure. These findings are grounded in $n=7$
specifications and should be read as observations within the
experiment rather than universal claims; the durable contribution is
the methodology, which produces harnesses as reusable artifacts that
feed live
tool environments~\citep{yao2024taubench} or simulated
agents~\citep{li2025simia}---closing the cold-start evaluation gap
for MCP-compatible tool suites.

\section*{Limitations}

\textbf{Ground truth reliability.}
The most significant limitation is reliance on LLM-generated ground
truth, which introduces systematic biases from the generating model.
The framework is best understood as a proxy evaluation tool that
identifies broad capability gaps and relative performance differences.

\textbf{Cross-call referential integrity.}
Mock outputs for sequential calls are currently generated independently,
meaning IDs or values may not align across dependent calls. A shared
state dictionary across a workflow would address this.

\textbf{Coverage ceiling.}
At 64 tools (Illustrator), coverage drops to 56\%. Targeted generation
strategies---such as iterative generation with coverage-aware tool
sampling---would be needed for larger specifications.

\textbf{Specification scope.}
Cross-domain harness generation and analysis spanning multiple MCP
specifications in a single workflow is structurally straightforward
within the existing pipeline as well, but is outside the scope of
this evaluation and remains a rich direction for future work.
Extension to other spec formats (OpenAPI, gRPC, function-calling
schemas) is similarly structurally straightforward---the same name +
typed-parameter scaffolding is present---and is a natural next step
for evaluating the framework's generality.

\textbf{Complexity stratification.}
The simple/complex distinction uses prompt framing rather than structural
enforcement. Post-generation filters based on structural complexity
metrics would improve stratification.

\textbf{Experimental scope.}
Seven MCP specifications and a single generation model (Gemini 2.5 Flash Lite)
cannot establish universal claims. Multi-turn evaluation records are
limited ($n\!=\!54$), with expansion heavily skewed toward complex
scenarios (30.8\% expansion rate) versus simple scenarios (2.8\%),
reflecting the stage's dependence on sufficient workflow substance to
generate meaningful follow-up turns. This constrains statistical power
for multi-turn findings.

\textbf{LLM-as-judge circularity.}
Both generation and quality verification rely on LLMs, raising the
concern that scenarios that ``look good to an LLM'' score well
regardless of actual quality. We address this in two ways. First, an
evaluator--generator capability gap: the judge (Gemini 2.5 Flash) has
surplus capacity over the weaker generator (Gemini 2.5 Flash Lite),
consistent with teacher--student evaluation paradigms; empirically,
Section~\ref{sec:experiments} shows the judge discriminates
meaningful, domain-specific failure patterns (argument cascading,
semantic ambiguity in Redis) rather than producing uniformly high
scores. Second, an out-of-family replication: re-scoring the full
391-record corpus with \texttt{Qwen3.5-122B-A10B-FP8} (Alibaba family)
yields paired Pearson $r{\approx}0.79$ on tool-calling ($n{=}384$)
with no mean shift, preserves the MCP ranking ($\rho{=}0.86$), and
reproduces the dominant argument value-accuracy failure mode
bilaterally (Gemini 240 records, Qwen 3.5 252; 4--5$\times$ margin
over the next sub-dim under both judges). Coherence shows a
systematic stricter-judge shift ($\Delta\mu{\approx}-0.16$) with
moderate record-level correlation ($r{=}0.42$); absolute coherence
levels and MCP-level coherence rank-preservation should therefore be
read as judge-dependent. Full
agreement tables, per-MCP bootstrap CIs under both judges, and the
Bland--Altman analysis appear in
Appendix~\ref{app:judge-replication}. A systematic human evaluation
study correlating framework scores with human quality judgments
remains an important direction for future work.

\bibliography{references}

\appendix
\section{Extended Experimental Results}
\label{app:extended}

\subsection{Evaluation Scale}

Table~\ref{tab:scale} breaks down the 337 generated scenarios and 391
evaluation records by MCP specification.
Records exceed scenarios because multi-turn scenarios contribute one
record per turn (single-turn entry plus one multi-turn entry per
additional turn).

\begin{table}[H]
\centering
\small
\begin{tabular}{@{}lrr@{}}
\toprule
& \textbf{Scenarios} & \textbf{Records} \\
\midrule
Illustrator   &  27 &  36 \\
Selenium      &  34 &  47 \\
Redis         &  96 &  98 \\
Git           &  72 &  85 \\
Elasticsearch &  42 &  49 \\
Slack         &  33 &  35 \\
Filesystem    &  33 &  41 \\
\midrule
\textbf{Total} & 337 & 391 \\
\bottomrule
\end{tabular}
\caption{Evaluation scale by MCP specification.}
\label{tab:scale}
\end{table}

\subsection{Complexity Breakdown}

Table~\ref{tab:complexity} reports mean tool-calling scores split by
complexity tier (simple vs.\ complex) for each MCP.
The $\Delta$ column shows the score change from simple to complex; a
negative value indicates quality degradation under higher complexity.
Elasticsearch shows the largest degradation ($-$12.2pp), followed by
Git ($-$11.1pp). Selenium is notably stable, scoring slightly higher
on complex scenarios.

\begin{table}[H]
\centering
\small
\begin{tabular}{@{}lccc@{}}
\toprule
\textbf{MCP} & \textbf{Simple} & \textbf{Complex} & $\boldsymbol{\Delta}$ \\
\midrule
Illustrator   & 0.934 & 0.865 & $-$0.069 \\
Selenium      & 0.932 & 0.935 & +0.003 \\
Redis         & 0.986 & 0.943 & $-$0.043 \\
Git           & 0.910 & 0.799 & $-$0.111 \\
Elasticsearch & 0.987 & 0.865 & $-$0.122 \\
Slack         & 0.934 & 0.840 & $-$0.094 \\
Filesystem    & 0.925 & 0.834 & $-$0.091 \\
\bottomrule
\end{tabular}
\caption{Tool-calling scores by complexity tier and MCP.}
\label{tab:complexity}
\end{table}

\subsection{Coherence Sub-Dimensions}

Table~\ref{tab:coh-subdim} reports mean scores for each of the five
coherence sub-dimensions on the raw 1--3 scale used by the judge.
Conciseness is near-ceiling across all domains (mean 2.90, std 0.33),
indicating the pipeline reliably produces appropriately scoped
responses without unnecessary elaboration.
Completeness is the primary bottleneck (mean 2.04), driven by
responses that address the user's request but omit contextual detail
that a practitioner would expect---the most actionable target for
future pipeline improvements.

\begin{table}[H]
\centering
\small
\begin{tabular}{@{}lcc@{}}
\toprule
\textbf{Sub-dimension} & \textbf{Mean} & \textbf{Std.} \\
\midrule
Conciseness       & 2.90 & 0.33 \\
Context retention & 2.78 & 0.59 \\
Logical flow      & 2.64 & 0.75 \\
Topic relevance   & 2.47 & 0.85 \\
Completeness      & 2.04 & 0.93 \\
\bottomrule
\end{tabular}
\caption{Coherence sub-dimension scores (raw 1--3 scale).}
\label{tab:coh-subdim}
\end{table}

\subsection{Grounding Tiers}

Table~\ref{tab:confidence} defines the grounding tiers assigned
to each mock output at Stage~3.
The tier records how much reference material was available to the LLM
when generating the synthetic tool response: \texttt{high} when
concrete example outputs were present in the tool specification,
\texttt{medium} when similar tools provided transferable examples,
and \texttt{low} when generation relied on the parameter schema alone.
In the current experiments, all seven MCP specifications lack example
outputs, so all 871 mock calls are annotated as \texttt{low}.
The annotation is carried through to the harness artifact for downstream
filtering when specs with richer documentation are used.

\begin{table}[H]
\centering
\small
\begin{tabular}{@{}lp{4.5cm}@{}}
\toprule
\textbf{Tier} & \textbf{Condition} \\
\midrule
\texttt{high}   & Concrete example outputs exist \\
\texttt{medium} & Examples from similar functions \\
\texttt{low}    & No examples available \\
\bottomrule
\end{tabular}
\caption{Grounding tiers for mock output generation.}
\label{tab:confidence}
\end{table}

\subsection{Workflow Composition and Category Diversity}
\label{app:diversity}

Selenium generates the longest workflows (mean 9.71 calls, max 17)
reflecting the sequential nature of browser automation tasks.
Illustrator workflows are also long (mean 3.33, max 12), while
Redis (mean 1.27) and Slack (mean 1.39) tend toward atomic
single-tool scenarios. Category count (108 unique categories across 337
scenarios; 3.1 scenarios per category) scales with scenario count rather
than tool count.

\subsection{Failure Details}
\label{app:failures}

\paragraph{Failure concentration by MCP.}
Git exhibits the highest failure rate: 7\% of records score below
0.5, followed by Filesystem (5\%) and Redis (1\%); four MCPs have
zero failures. Git's failures are driven by argument correctness for
tools with complex parameter schemas.

\paragraph{Argument failure patterns.}
Among argument failures, \textit{value accuracy} is the dominant
sub-dimension (223 records), followed by \textit{relevancy} (44),
\textit{format compliance} (35), \textit{type compliance} (31),
\textit{completeness} (16), and \textit{name accuracy} (11).

\begin{table}[H]
\centering
\small
\begin{tabular}{@{}p{5.2cm}r@{}}
\toprule
\textbf{Issue Group} & \textbf{Count} \\
\midrule
Missing info / shallow response   & 373 \\
Off-topic or wrong focus          & 130 \\
Non-sequitur or topic shift       &  86 \\
Self-contradiction / hallucination &  85 \\
Ignores user constraints          &  54 \\
No shared context (eval artifact) &  44 \\
Unnecessary elaboration           &  35 \\
Lost thread / forgot prior info   &  13 \\
\bottomrule
\end{tabular}
\caption{Coherence issue taxonomy grouped from evaluator notes.}
\label{tab:coh-issues}
\end{table}

\subsection{Per-MCP Dimension Scores}

\begin{table}[H]
\centering
\small
\begin{tabular}{@{}lcccc@{}}
\toprule
\textbf{MCP} & \textbf{Usage} & \textbf{Select} & \textbf{Order} & \textbf{Args} \\
\midrule
Illustrator   & 0.996 & 0.835 & 0.800 & 0.872 \\
Selenium      & 0.965 & 0.891 & 0.989 & 0.890 \\
Redis         & 1.000 & 0.954 & 0.823 & 0.952 \\
Git           & 0.988 & 0.802 & 0.761 & 0.780 \\
Elasticsearch & 1.000 & 0.868 & 0.790 & 0.907 \\
Slack         & 1.000 & 0.867 & 0.576 & 0.865 \\
Filesystem    & 1.000 & 0.791 & 0.659 & 0.894 \\
\bottomrule
\end{tabular}
\caption{Mean TC dimension scores by MCP (overall). Selenium's
ordering (0.989) is the highest of any MCP, despite having the longest
workflows. Git's arguments (0.780) are weakest, consistent with its
high parameter density.}
\label{tab:mcp-dims}
\end{table}

\subsection{Out-of-Family Judge Replication}
\label{app:judge-replication}

To assess whether reported scores are artifacts of within-family judge
circularity, the full 391-record evaluation corpus was re-scored with
an out-of-family judge: Qwen3.5-122B-A10B-FP8
(Alibaba family; vLLM-hosted, FP8 quantization). Each record's existing
chat transcript was replayed against the alternate judge with only the
evaluator model swapped; the generation pipeline and rubric are
unchanged. After joining on the composite key (MCP,
\texttt{conversation\_id}), 384 records have paired tool-calling
scores and 380 have paired coherence scores under both judges; all
paired statistics below are computed on these sets.

\paragraph{Per-MCP means with bootstrap CIs.}
Table~\ref{tab:judge-permcp} reports per-MCP means and 95\% percentile
bootstrap CIs ($B{=}2{,}000$) under both judges. Tool-calling CIs
overlap for every MCP. Coherence CIs are disjoint for all seven MCPs
(the Qwen 3.5 judge is systematically lower), and the identity of
the worst-coherence MCP differs across judges (Git under Gemini,
Illustrator under Qwen 3.5).

\begin{table}[H]
\centering
\small
\resizebox{\columnwidth}{!}{%
\begin{tabular}{@{}lrcccc@{}}
\toprule
\textbf{MCP} & \textbf{n} & \textbf{TC Gem.} & \textbf{TC Qwen} & \textbf{Coh Gem.} & \textbf{Coh Qwen} \\
\midrule
Redis         & 98 & 0.966 & 0.964 & 0.902 & 0.766 \\
Elasticsearch & 49 & 0.930 & 0.935 & 0.902 & 0.744 \\
Illustrator   & 36 & 0.898 & 0.821 & 0.855 & 0.604 \\
Slack         & 35 & 0.886 & 0.884 & 0.938 & 0.707 \\
Filesystem    & 41 & 0.876 & 0.871 & 0.825 & 0.701 \\
Git           & 85 & 0.857 & 0.858 & 0.757 & 0.647 \\
Selenium      & 47 & 0.935 & 0.919 & 0.850 & 0.655 \\
\bottomrule
\end{tabular}}%
\caption{Per-MCP means under the Gemini judge
(\texttt{gemini-2.5-flash-lite}) and the Qwen 3.5 judge
(\texttt{Qwen3.5-122B-A10B-FP8}). $n$ counts come from the (MCP,
\texttt{conversation\_id}) join, which preserves all records;
Gemini means match those reported in Table~\ref{tab:quality-mcp}.
95\% bootstrap CIs (omitted for space) overlap for every MCP on TC;
coherence CIs are disjoint for all 7 MCPs. The bottom-coherence MCP
differs across judges (Git under Gemini, Illustrator under Qwen 3.5).}
\label{tab:judge-permcp}
\end{table}

MCP-level Spearman rank correlations across the seven MCPs are
$\rho{=}0.86$ for tool-calling and $\rho{=}0.46$ for coherence. The
top tool-calling MCP (Redis) is preserved under both judges; the
bottom-coherence MCP differs (Git under Gemini, Illustrator under
Qwen 3.5).

\paragraph{Record-level agreement.}
Table~\ref{tab:judge-agree} reports paired-record agreement on the
aggregate per-aspect unsupervised score
(\texttt{agg\_unsupervised\_score}, the metric reported throughout the
paper). Tool-calling agreement is strong; coherence is moderate.

\begin{table}[H]
\centering
\small
\begin{tabular}{@{}lrcccc@{}}
\toprule
\textbf{Aspect} & \textbf{n} & \textbf{Pearson r} & \textbf{Spearman $\rho$} & \textbf{MAE} & \textbf{$\pm$0.1} \\
\midrule
Tool-calling & 384 & 0.790 & 0.844 & 0.046 & 84\% \\
Coherence    & 380 & 0.420 & 0.438 & 0.180 & 34\% \\
\bottomrule
\end{tabular}
\caption{Record-level agreement between judges, computed on records
keyed by (MCP, \texttt{conversation\_id}) to avoid scenario-id
collisions across MCPs. Bootstrap 95\% CIs ($B{=}2{,}000$) omitted
for space. MAE and the $\pm 0.1$ agreement rate reproduce the
within-tolerance qualitative pattern of the original analysis.}
\label{tab:judge-agree}
\end{table}

\paragraph{Systematic shift.}
The mean signed difference (Qwen 3.5 $-$ Gemini) is
$\Delta\mu_\textrm{TC}{\approx}0$ (95\% CI $[-0.009,+0.008]$;
balanced sign-test) and $\Delta\mu_\textrm{Coh}{\approx}-0.16$
(95\% CI $[-0.173,-0.138]$; the majority of non-tied records score
lower under the Qwen 3.5 judge).
Tool-calling shows no systematic bias; coherence is systematically
stricter under the out-of-family judge. Without a third judge or a
human anchor, neither judge can be designated ``correct'' on
coherence; absolute coherence levels are therefore reported as
judge-dependent.

\paragraph{Failure-mode taxonomy replication.}
Table~\ref{tab:judge-argfail} reports record counts where each
argument sub-dimension scored below 1.0 under each judge. Both judges
identify \textit{value\_accuracy} as the dominant argument failure
mode by a 4--5$\times$ margin over the next-highest sub-dimension,
replicating the central qualitative finding of
Section~\ref{sec:experiments}; counts are within 5\% across judges.

\begin{table}[H]
\centering
\small
\begin{tabular}{@{}lrr@{}}
\toprule
\textbf{Sub-dimension} & \textbf{Gemini} & \textbf{Qwen 3.5} \\
\midrule
\textbf{value\_accuracy} & \textbf{240} & \textbf{252} \\
relevancy          & 53 & 54 \\
format\_compliance & 46 & 58 \\
type\_compliance   & 46 & 42 \\
completeness       & 23 & 33 \\
name\_accuracy     & 16 & 21 \\
\bottomrule
\end{tabular}
\caption{Argument sub-dimension failure counts (records with score
$<1.0$) under each judge. Value-accuracy dominance is preserved
bilaterally; counts agree within 5\% across judges.}
\label{tab:judge-argfail}
\end{table}

\begin{figure}[H]
\centering
\includegraphics[width=\columnwidth]{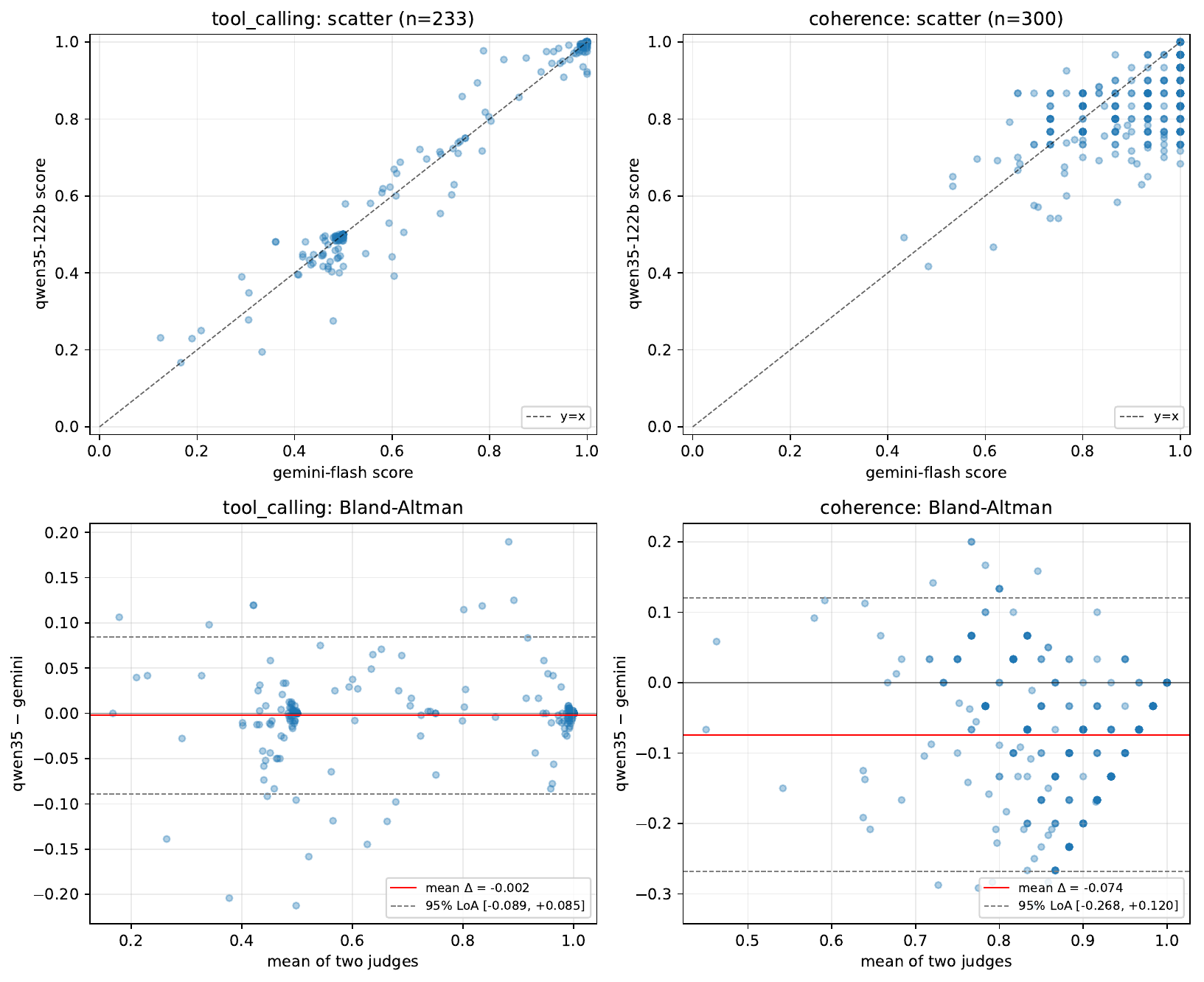}
\caption{Record-level agreement between the Gemini judge
(\texttt{gemini-2.5-flash-lite}) and the Qwen 3.5 judge
(\texttt{Qwen3.5-122B-A10B-FP8}). Top: scatter (Qwen 3.5 vs.\ Gemini); the
TC cloud (left) lies on $y{=}x$, the coherence cloud (right) is offset
below. Bottom: Bland--Altman; TC differences are centered on zero,
coherence differences are shifted ${\approx}{-}0.138$.}
\label{fig:judge-bland-altman}
\end{figure}

\paragraph{Visual diagnostics.}
Figure~\ref{fig:judge-bland-altman} shows scatter (top row) and
Bland--Altman (bottom row) plots for tool-calling and coherence. The
TC scatter clusters tightly along $y{=}x$ and the TC Bland--Altman
limits-of-agreement bracket zero symmetrically; the coherence scatter
shows a visible offset below $y{=}x$ and the coherence Bland--Altman
LoA is shifted ${\approx}{-}0.14$ below zero with wider spread.

\paragraph{Interpretation.}
Per-MCP TC means, the MCP TC ranking, and the dominant
argument-value-accuracy failure mode are robust to the judge swap.
Absolute coherence levels and MCP-level coherence rank-preservation
($\rho_{\textrm{Coh}}{=}0.46$) are judge-dependent. The
worst-coherence MCP differs across judges (Git under Gemini,
Illustrator under Qwen 3.5), reflecting the broader judge-dependence
of coherence scoring rather than a contradiction in the underlying
data.

\begin{table}[H]
\centering
\small
\resizebox{\columnwidth}{!}{%
\begin{tabular}{@{}lrrrrrcc@{}}
\toprule
\textbf{MCP} & \textbf{$|T|$} & \textbf{$\bar{p}$} & \textbf{$p_{\max}$} & \textbf{Opt\%} & \textbf{$\bar{w}$} & \textbf{TC} & \textbf{Coh} \\
\midrule
Illustrator   & 64 & 3.6 & 12 & 74 & 3.3 & 0.898 & 0.855 \\
Selenium      & 56 & 1.8 &  5 & 26 & 9.7 & 0.935 & 0.850 \\
Redis         & 47 & 2.1 &  6 & 34 & 1.3 & 0.966 & 0.902 \\
Git           & 33 & 11.2 & 19 & 95 & 2.2 & 0.857 & 0.757 \\
Elasticsearch & 20 & 1.8 &  8 & 33 & 1.9 & 0.930 & 0.902 \\
Slack         & 16 & 2.2 &  6 & 51 & 1.4 & 0.886 & 0.938 \\
Filesystem    & 14 & 1.8 &  3 & 32 & 2.0 & 0.876 & 0.825 \\
\bottomrule
\end{tabular}}%
\caption{Schema complexity profile. $|T|$ = tools, $\bar{p}$ = mean
params/tool, $p_{\max}$ = max params, Opt\% = optional parameter ratio,
$\bar{w}$ = mean workflow length. Per-MCP correlations with mean TC:
average parameters per tool $r=-0.60$, optional ratio $r=-0.66$,
tool count $r=+0.40$ (positive but smaller). Tool-level
disaggregation ($n=222$) confirms the schema-complexity direction
with $p<0.001$: average parameters $r=-0.29$, optional ratio
$r=-0.30$, both relative to mean unsupervised TC.}
\label{tab:schema-complexity}
\end{table}

\section{Aggregation Sensitivity}
\label{app:aggregation-sensitivity}

The reported tool-calling scores combine the four top-level dimensions
via arithmetic mean. Table~\ref{tab:aggregation-sensitivity}
re-aggregates the same per-turn dimension scores under harmonic mean
and minimum-across-dimensions ($n=385$; six coherence-only records
omitted). Corpus-wide means shift substantially
(0.911 / 0.851 / 0.781), but per-MCP rankings are highly stable:
Spearman $\rho = 0.964$ (Arith.~vs Harm.), $0.964$ (Arith.~vs Min.),
$0.929$ (Harm.~vs Min.). The only swap is Filesystem
$\leftrightarrow$ Git near the bottom; the top-3 ordering is identical
under all three rules. The schema-complexity correlation
(Section~\ref{sec:experiments}) holds in direction throughout:
Pearson $r = -0.60$ (Arith.), $-0.48$ (Harm.), $-0.50$ (Min.) across
the seven MCPs, and localizes to the argument sub-dimension
($r=-0.86$ versus $|r|\leq 0.46$ for the other three). Argument
correctness is also the dominant failure mode at the input level
(60.5\% of 448 turns score below 1.0, vs.~29.1\% for ordering, 28.6\%
for selection, 1.3\% for usage)---a turn-level fact unaffected by the
outer aggregation choice.

\begin{table}[t]
\centering
\small
\begin{tabular}{@{}lrrrr@{}}
\toprule
\textbf{MCP} & \textbf{$|T|$} & \textbf{Arith.} & \textbf{Harm.} & \textbf{Min.} \\
\midrule
Redis         & 47 & 0.966 & 0.942 & 0.905 \\
Selenium      & 56 & 0.935 & 0.909 & 0.814 \\
Elasticsearch & 20 & 0.930 & 0.890 & 0.821 \\
Illustrator   & 64 & 0.898 & 0.839 & 0.745 \\
Slack         & 16 & 0.886 & 0.813 & 0.737 \\
Filesystem    & 14 & 0.876 & 0.746 & 0.680 \\
Git           & 33 & 0.857 & 0.763 & 0.677 \\
\midrule
\textbf{Corpus} &  & \textbf{0.911} & \textbf{0.851} & \textbf{0.781} \\
\bottomrule
\end{tabular}
\caption{Per-MCP TC means under arithmetic, harmonic, and
minimum aggregation across the four dimensions ($n=385$).}
\label{tab:aggregation-sensitivity}
\end{table}

\section{Novelty Summary}
\label{app:novelty}

Table~\ref{tab:novelty} situates the contributions against prior
literature across two gaps: the data generation gap (rows 1--3) and
the metric decomposition gap (rows 4--6).

\begin{table*}[t]
\centering
\footnotesize
\renewcommand{\arraystretch}{1.3}
\resizebox{\textwidth}{!}{%
\begin{tabular}{@{}p{4.6cm}p{5.8cm}p{5.2cm}@{}}
\toprule
\textbf{Contribution} & \textbf{Prior Literature} & \textbf{This Work} \\
\midrule
Benchmark construction
  & Manual curation; or LLM synthesis for training data
    \citep{verma2025fabric,castellani2025synthtools,liu2024apigen}
  & Automated \textit{evaluation} harnesses from MCP specs, with
    held-out oracles and mock outputs \\
Mock output generation
  & Live API execution~\citep{qin2023toolllm}; or neural simulator
    trained on large corpora \citep{gtm2025,castellani2025synthtools}
  & LLM-prompted generation from spec alone; no live API or trained simulator \\
Multi-turn grounding
  & Context-free follow-ups
  & Mock-data-conditioned turn generation \\
\midrule
Tool-calling granularity
  & Name + parameter match; argument-flow via
    DAG~\citep{maekawa2025funcbenchgen}
  & 4-dimension decomposition (usage/selection/ordering/arguments) \\
Argument scoring strategy
  & Arithmetic mean of parameter scores
  & Arithmetic mean with judge-enforced cascading penalties
    on critical errors \\
Failure taxonomy
  & Pass/fail
  & 6-category failure type classification \\
\bottomrule
\end{tabular}
}
\caption{Contributions vs.\ prior literature, organized by thematic gap.}
\label{tab:novelty}
\end{table*}

\section{Pipeline Prompts}
\label{app:prompts}

This section reproduces the instructional content of the LLM prompts
used at each pipeline stage. JSON output templates and tool-spec
payloads are abbreviated for space.

\subsection{Stage 1: Tool Interpretation}

The tool interpreter receives a single MCP tool specification and
produces a structured semantic explanation, requested as a JSON
object with five named fields.

\begin{small}
\begin{verbatim}
I have a tool that can be called by an agent,
and I could use help understanding what it
does and what it is helpful for.

Tool info:
```json
{tool_info}
```

I need a json in the following format that
can help me thoroughly understand what the
tool is capable of, especially in an
enterprise context. Keep the explanations
grounded within the tool info.

{
  "tool_name": <Tool name here, as given>,
  "what_it_does": <Complete explanation of
      the tool's functionality and what it
      aims to do>,
  "what_it_needs": <What parameters the tool
      needs and how they should be formatted>,
  "why_its_used": <Reasons an agent would
      call this tool; potential use cases>,
  "enterprise_context": <Tags for what aspect
      of an enterprise this could help with>
}
\end{verbatim}
\end{small}

\subsection{Stage 2: Scenario Generation}

Two prompts generate scenarios at different complexity levels from
the enriched tool summaries. Both target an ``agentic chatbot for
enterprise use cases'' and request scenarios organized by category,
each containing an exact \code{agent\_workflow} of function calls
with parameters.

\paragraph{Simple scenarios.}
\begin{small}
\begin{verbatim}
I'm building an agentic chatbot for enterprise
use cases. Based on the available tool
capabilities below, generate realistic,
straightforward, and commonplace scenarios
organized by category that showcase how
employees would use this chatbot for everyday
tasks. These examples would not require too
many tool calls -- they'll be smaller and
more precise.

Available Tool Capabilities:
{tool_summary}

For each scenario, include the exact function
calls the agent would make using the available
tools.

[JSON format omitted: categories[].scenarios[]
 with title, prompt, agent_workflow[], novelty
 _reason, agent_followup]

Make sure:
 1. Use actual tool names from the available
    capabilities
 2. Function names and parameters are
    structured separately with realistic
    values adhering to the parameter schema
 3. Workflows show logical progression
 4. Scenarios are practical and commonly
    encountered
 5. Agent workflows do the necessary context
    management & tool calls to identify how
    parameters are selected
 6. Provide meaningful agent_followup content
    that makes sense within the context of
    the scenario.
\end{verbatim}
\end{small}

\paragraph{Complex scenarios.}
The complex prompt mirrors the simple prompt but replaces
``straightforward, and commonplace'' with ``novel, and complex'',
asks for advanced and creative tool usage, and adds two ``Make
sure'' items: \textit{each scenario demonstrates complex,
multi-step processes} and \textit{creative combinations of tools
that unlock new capabilities}.

\paragraph{Coverage hint and follow-up.}
A coverage suffix appended to the initial prompt instructs:
\textit{``IMPORTANT: Ensure broad coverage across ALL available
tools. Every tool listed above should appear in at least one
scenario's agent\_workflow. There are \{N\} tools total --- design
scenarios that collectively exercise all of them.''} If tools remain
uncovered after the first round, a follow-up prompt requests
additional scenarios for the named uncovered tools, allowing
combination with previously covered tools in multi-step workflows.

\subsection{Stage 3: Mock Output Generation}

\begin{small}
\begin{verbatim}
You are a Mock Tool Output Generator for
synthetic agent workflow data. Your task is
to generate realistic mock tool outputs that
complete synthetic scenarios.

You will be given an initial prompt (and
maybe a description of what the aim of the
prompt is & why the scenario is of interest).
Then, you will be given an agent workflow.
This will detail:
  (1) A function call w/ parameters and
  (2) A quick explanation of what the tool
      does.

Finally, you will have some example function
calls with their respective outputs OR a JSON
schema object describing the output
structure. Use this to guide formatting for
the final mock tool output.

[JSON output: mock_workflow[] with function
 _name, parameters, quick_explanation,
 mock_output, confidence; plus expected
 _response that references specific mock data]

### CONFIDENCE LEVEL GUIDELINES
"high":   concrete example for THIS specific
          function was provided
"medium": no example for this function, but
          similar functions have examples
"low":    no example output provided for
          this function

### CRITICAL INSTRUCTIONS
 1. Concrete Data: replace placeholders (e.g.
    "{user_id}" or "XYZ") with realistic,
    specific values.
 2. Realism & Diversity: reflect how a real
    system would respond; incorporate diverse
    names, global locations, and varied data
    points.
 3. Formatting: strictly adhere to provided
    reference examples or JSON schema.
 4. Expected response references concrete
    mock data (names, IDs, counts, statuses,
    dates) and reflects the full workflow,
    not just the last call.
\end{verbatim}
\end{small}

\subsection{Stage 4: Multi-Turn Expansion}

The multi-turn expansion prompt is constructed in parts: a preamble,
scenario and tool-result context, key principles, task instructions,
worked examples of good vs.\ bad turn splitting, quality
guidelines, and a JSON output template. The principle set differs
between \textit{mock-data-grounded} mode (when synthetic outputs
exist) and \textit{standard} mode.

\begin{small}
\begin{verbatim}
You are a Natural Conversation Flow Analyzer
for enterprise agent interactions. Your task
is to take a scenario and intelligently break
it into natural conversation turns that
reflect how real employees would interact
with an agent.

# Key Principles (mock-data-grounded mode)
 1. Use Available Tool Results to inform
    realistic follow-up questions and workflow
    decisions.
 2. Data-Driven Breakpoints from the actual
    data returned by tools.
 3. Realistic User Reactions to the specific
    data shown.
 4. Progressive Data Exploration: each turn
    builds on prior tool outputs.
 5. Split at Phase Boundaries when distinct
    phases exist (e.g., information-gathering
    then acting on it).
 6. Group Related Operations Within a Phase:
    batch repeated operations (e.g., 3
    lookups) into a SINGLE turn.
 7. Self-Contained Turns: agent_followup
    reports concrete results, NOT narration of
    future actions.
 8. Complete the Full Workflow: every tool
    call from the initial workflow appears in
    exactly one turn.
 9. CRITICAL CONSTRAINT - Use Only Existing
    Functions: same function names and
    parameter structures as the initial
    workflow.

[Worked examples follow: BAD over-granular
 splitting, BAD no-splitting, GOOD batched-
 within-phases. JSON format omitted: turns[]
 with title, prompt, agent_workflow[]
 including mock_output and confidence,
 novelty_reason, agent_followup]
\end{verbatim}
\end{small}

In \textit{standard} mode (no mock outputs available), the
principles drop the data-driven breakpoint and tool-result
references, and the JSON template omits the \code{mock\_output} and
\code{confidence} fields. The phase-boundary, batching,
self-contained-turn, and function-reuse constraints are preserved
across both modes.

\section{Evaluation Prompts}
\label{app:eval-prompts}

The quality assessment framework (Section~\ref{sec:scoring}) uses
two LLM-as-judge prompts: one for tool-calling correctness and one
for conversational coherence. Both operate in unsupervised mode
(no reference answer). Tables~\ref{tab:tc-rubric}
and~\ref{tab:coh-dims} detail the aspects measured, their
definitions, and scoring scales.

\subsection{Tool-Calling Evaluation}

The tool-calling prompt evaluates across four dimensions, each with
scored sub-dimensions on a 0--10 scale. Sub-scores are normalized to
0--1; the selection, ordering, and argument dimension scores are the
arithmetic mean of their sub-scores, while the usage dimension score
is taken directly from its necessity sub-score. The four top-level
dimension scores are then combined via arithmetic mean. Cascading
penalties on argument sub-scores are enforced through prompt
instructions to the judge (Table~\ref{tab:tc-rubric}, footnote).

\begin{table*}[t]
\centering
\small
\begin{tabular}{@{}llp{7.5cm}c@{}}
\toprule
\textbf{Dimension} & \textbf{Sub-dimension} & \textbf{Definition} & \textbf{Scale} \\
\midrule
\multirow{2}{*}{Usage} & Necessity & Was a tool actually needed, or could the assistant answer directly? & 0--10 \\
 & Overuse detection & Are there redundant or unnecessary tool calls? & 0--10 \\
\midrule
\multirow{3}{*}{Selection} & Correctness & Do the selected tools match the task described by the user? & 0--10 \\
 & Specificity & Was the most specific tool chosen when alternatives exist? & 0--10 \\
 & Completeness & Are all tools needed to fully address the query called? & 0--10 \\
\midrule
\multirow{3}{*}{Ordering} & Sequence logic & Is the execution order logical; do later calls build on earlier ones? & 0--10 \\
 & Dependency handling & Are inter-tool dependencies respected (output $\rightarrow$ input)? & 0--10 \\
 & Execution efficiency & Could reordering improve efficiency? & 0--10 \\
\midrule
\multirow{6}{*}{Arguments} & Completeness & Are all required parameters provided? & 0--10 \\
 & Name accuracy & Do parameter names match schemas exactly (case-sensitive)? & 0--10 \\
 & Value accuracy & Are values correct and grounded in the user query or prior tool outputs? & 0--10 \\
 & Type compliance & Do parameter values match expected data types? & 0--10 \\
 & Format compliance & Do values follow expected formats (dates, enums, patterns)? & 0--10 \\
 & Relevancy & Are there any extra or invalid parameters not in the schema? & 0--10 \\
\bottomrule
\end{tabular}
\caption{Tool-calling evaluation dimensions and sub-dimensions. Ordering is marked
not applicable for single tool calls and excluded from aggregation.
\textbf{Cascading rules} (enforced via prompt instructions to the
judge): when the judge identifies a wrong
parameter name (score~$\leq$~2), it is instructed to assign
near-zero scores to the dependent argument sub-dimensions (value,
type, format); a missing required parameter triggers the same
cascade; a wrong value (score~$\leq$~3) cascades to type, format,
and relevancy. Values from prior tool outputs in chained calls are
not penalized.}
\label{tab:tc-rubric}
\end{table*}

\subsection{Coherence Evaluation}

The coherence prompt evaluates across five dimensions on a 1--3
scale, normalized to 0--1 and aggregated via arithmetic mean.
Each dimension checks for specific failure manifestations.

\begin{table*}[t]
\centering
\small
\begin{tabular}{@{}lp{6.5cm}p{5.5cm}@{}}
\toprule
\textbf{Dimension} & \textbf{Definition} & \textbf{Failure manifestations} \\
\midrule
Context retention & How well the response maintains and uses information from conversation history &
\code{asks\_again}, \code{forgets\_preferences}, \code{pronoun\_confusion},
\code{contradicts\_self}, \code{loses\_thread}, \code{ignores\_corrections} \\
\midrule
Logical flow & How well the response follows logically from previous turns and maintains coherent progression &
\code{topic\_shift}, \code{non\_sequitur}, \code{poor\_transitions},
\code{breaks\_causality}, \code{temporal\_confusion} \\
\midrule
Completeness & How thoroughly the response addresses all parts of the user's query &
\code{cuts\_off}, \code{partial\_answer}, \code{too\_shallow},
\code{ignores\_constraints}, \code{missing\_key\_info}, \code{no\_actionable\_advice} \\
\midrule
Conciseness & How efficiently the response communicates without unnecessary repetition &
\code{repeats\_directly}, \code{rephrases\_same\_point}, \code{too\_much\_fluff},
\code{over\_explains}, \code{excessive\_caution} \\
\midrule
Topic relevance & How well the response stays focused on the user's query and conversation topic &
\code{complete\_topic\_shift}, \code{misses\_main\_point}, \code{too\_generic},
\code{hallucination}, \code{wrong\_question} \\
\bottomrule
\end{tabular}
\caption{Coherence evaluation dimensions. Each dimension is scored on a 1--3 scale:
\textbf{Good}~(3) = no manifestations detected;
\textbf{Adequate}~(2) = 1--2 minor manifestations;
\textbf{Poor}~(1) = 3+ manifestations or critical failures.
Context retention is optional (excluded when no conversation history exists).}
\label{tab:coh-dims}
\end{table*}

\section{Structured Output Schemas}
\label{app:schemas}

\begin{figure*}[t]
\centering
\begin{tikzpicture}[
  schemabox/.style={
    draw=black!70,
    rounded corners=6pt,
    fill=white,
    minimum width=2.8cm,
    align=left,
    inner sep=5pt,
    font=\small
  },
  stagelabel/.style={
    font=\scriptsize\bfseries,
    text=black!50,
    anchor=south
  },
  arr/.style={
    -{Stealth[length=5pt]},
    thick,
    black!50
  },
  inheritarr/.style={
    -{Stealth[length=5pt, open]},
    thick,
    black!40,
    dashed
  },
  stageheader/.style={
    font=\footnotesize\bfseries,
    text=black!40
  }
]


\node[schemabox] (toolinfo) {
  \textbf{\texttt{ToolInfo}}\\[2pt]
  \texttt{\scriptsize name: str}\\
  \texttt{\scriptsize description: str}\\
  \texttt{\scriptsize input\_schema: dict}\\
  \texttt{\scriptsize annotations: dict}
};
\node[stagelabel] at (toolinfo.north) {Input};

\node[schemabox, right=0.8cm of toolinfo] (toolexp) {
  \textbf{\texttt{ToolExplanation}}\\[2pt]
  \texttt{\scriptsize tool\_name: str}\\
  \texttt{\scriptsize what\_it\_does: str}\\
  \texttt{\scriptsize why\_its\_used: str}\\
  \texttt{\scriptsize what\_it\_needs: str}\\
  \texttt{\scriptsize enterprise\_context: str}
};

\node[schemabox, right=0.8cm of toolexp] (scenario) {
  \textbf{\texttt{Scenario}}\\[2pt]
  \texttt{\scriptsize title: str}\\
  \texttt{\scriptsize prompt: str}\\
  \texttt{\scriptsize agent\_workflow:}\\
  \texttt{\scriptsize \quad list[AgentCall]}\\
  \texttt{\scriptsize novelty\_reason: str}\\
  \texttt{\scriptsize agent\_followup: str}
};

\draw[arr] (toolinfo) -- (toolexp);
\draw[arr] (toolexp) -- (scenario);

\node[stageheader, above=0.3cm of toolinfo.north west, anchor=south west]
  {Stage 1: Interpret};
\node[stageheader, above=0.3cm of scenario.north west, anchor=south west]
  {Stage 2: Generate};


\node[schemabox, below=1.4cm of toolinfo, fill=black!4] (agentcall) {
  \textbf{\texttt{AgentCall}}\\[2pt]
  \texttt{\scriptsize function\_name: str}\\
  \texttt{\scriptsize parameters: dict}\\
  \texttt{\scriptsize quick\_explanation: str}
};

\node[schemabox, right=0.8cm of agentcall, fill=black!4] (mockoutput) {
  \textbf{\texttt{MockOutput}}\\
  {\scriptsize\textit{extends AgentCall}}\\[2pt]
  \texttt{\scriptsize mock\_output: str|dict}\\
  \texttt{\scriptsize confidence: Enum}\\
  \texttt{\scriptsize \quad high|medium|low}
};

\node[schemabox, right=0.8cm of mockoutput] (mockwf) {
  \textbf{\texttt{MockWorkflow}}\\[2pt]
  \texttt{\scriptsize mock\_workflow:}\\
  \texttt{\scriptsize \quad list[MockOutput]}\\
  \texttt{\scriptsize expected\_response: str}
};

\node[schemabox, right=0.8cm of mockwf] (multiturn) {
  \textbf{\texttt{MockMultiTurn-}}\\
  \textbf{\texttt{Scenario}}\\[2pt]
  \texttt{\scriptsize turns: list[Turn]}\\[2pt]
  \textit{\scriptsize each turn:}\\
  \texttt{\scriptsize \quad prompt: str}\\
  \texttt{\scriptsize \quad agent\_workflow:}\\
  \texttt{\scriptsize \quad\quad list[MockOutput]}\\
  \texttt{\scriptsize \quad agent\_followup: str}
};

\draw[inheritarr] (agentcall) -- (mockoutput);
\draw[arr] (mockoutput) -- (mockwf);
\draw[arr] (mockwf) -- (multiturn);

\draw[inheritarr] (agentcall.north) -- ++(0,0.5) -| (scenario.south);

\node[stageheader, above=0.3cm of mockoutput.north west, anchor=south west]
  {Stage 3: Mock};
\node[stageheader, above=0.3cm of multiturn.north west, anchor=south west]
  {Stage 4: Expand};

\end{tikzpicture}
\caption{Structured output schemas across the four pipeline stages.
Solid arrows indicate data flow; dashed arrows indicate schema
inheritance (\code{MockOutput} extends \code{AgentCall}, which is
referenced by \code{Scenario}).}
\label{fig:schemas}
\end{figure*}
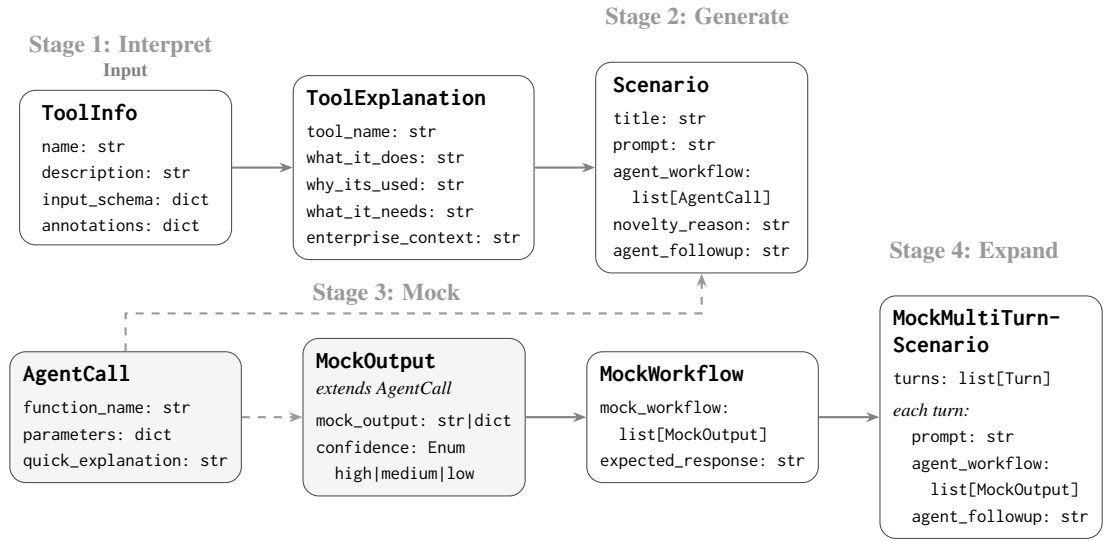

Each pipeline stage uses Pydantic models as structured output schemas,
constraining the LLM to produce well-formed JSON at every step.

\section{Score Distributions and Per-Domain Analysis}
\label{app:figures}

\subsection{Overall Score Distributions}

\begin{figure*}[t]
\centering
\begin{subfigure}[t]{0.49\textwidth}
  \centering
  \includegraphics[width=\textwidth]{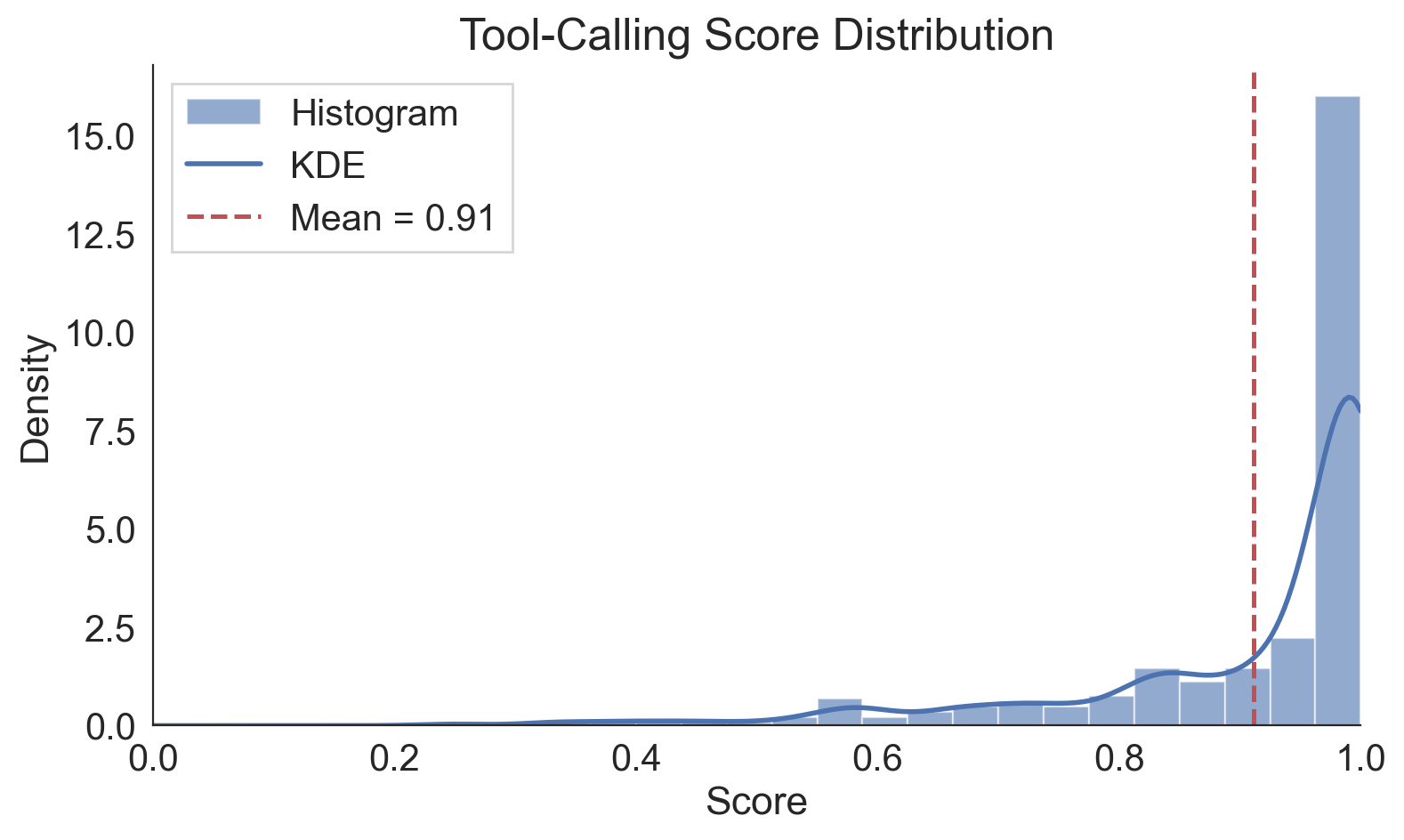}
  \caption{Tool-calling score distribution with KDE.}
  \label{fig:tc-dist}
\end{subfigure}
\hfill
\begin{subfigure}[t]{0.49\textwidth}
  \centering
  \includegraphics[width=\textwidth]{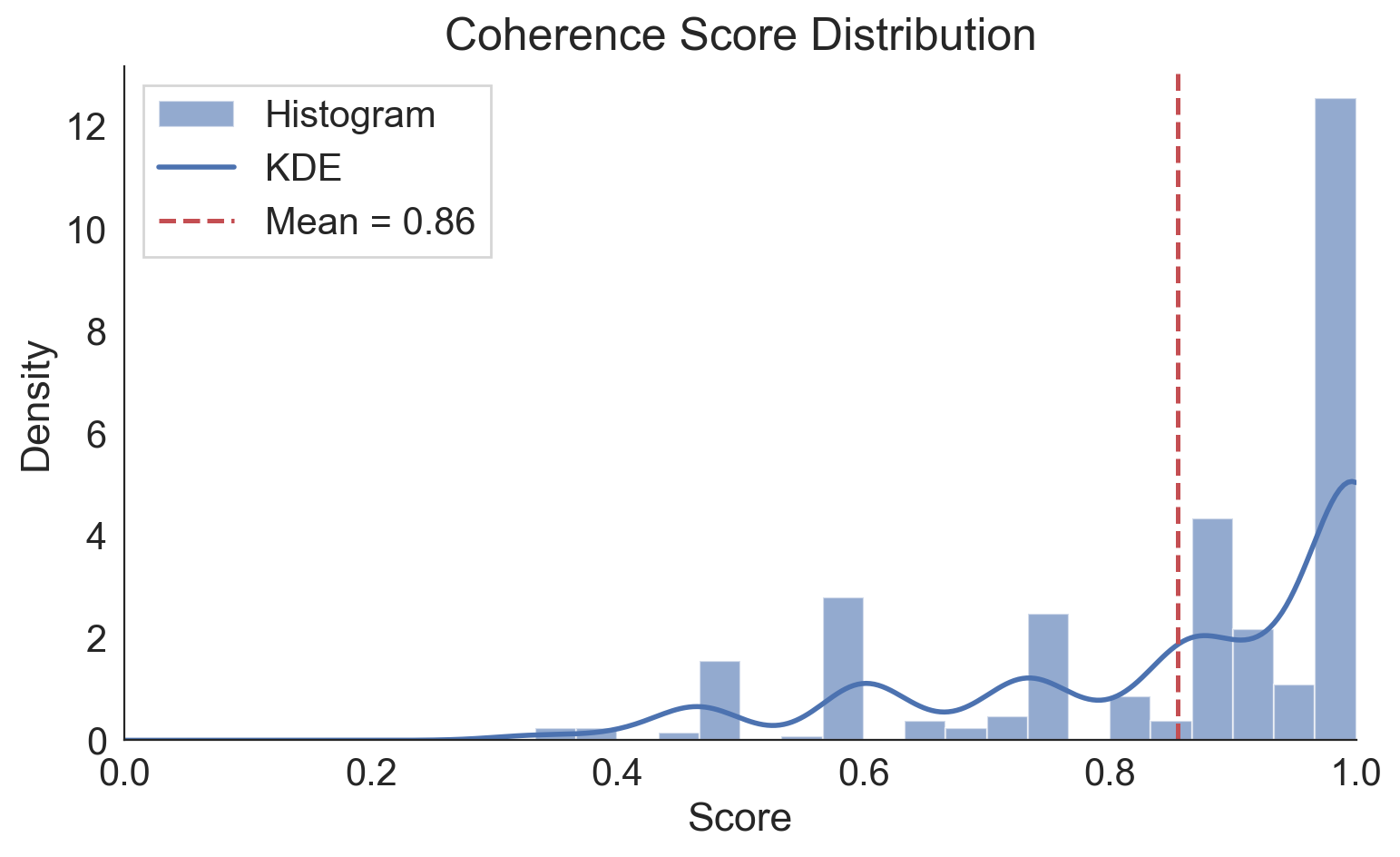}
  \caption{Coherence score distribution with KDE.}
  \label{fig:coh-dist}
\end{subfigure}
\caption{Score distributions across all 391 evaluation records.}
\label{fig:distributions}
\end{figure*}

Figure~\ref{fig:distributions} shows the distribution of tool-calling
and coherence scores across all 391 evaluation records. The
tool-calling distribution is concentrated in the upper range (31.7\%
perfect scores, 2.3\% below 0.5).
The coherence distribution is left-skewed with a mode near 0.9.

\subsection{Per-Domain Analysis}

\begin{figure*}[t]
\centering
\begin{subfigure}[t]{0.46\textwidth}
  \centering
  \includegraphics[width=\textwidth]{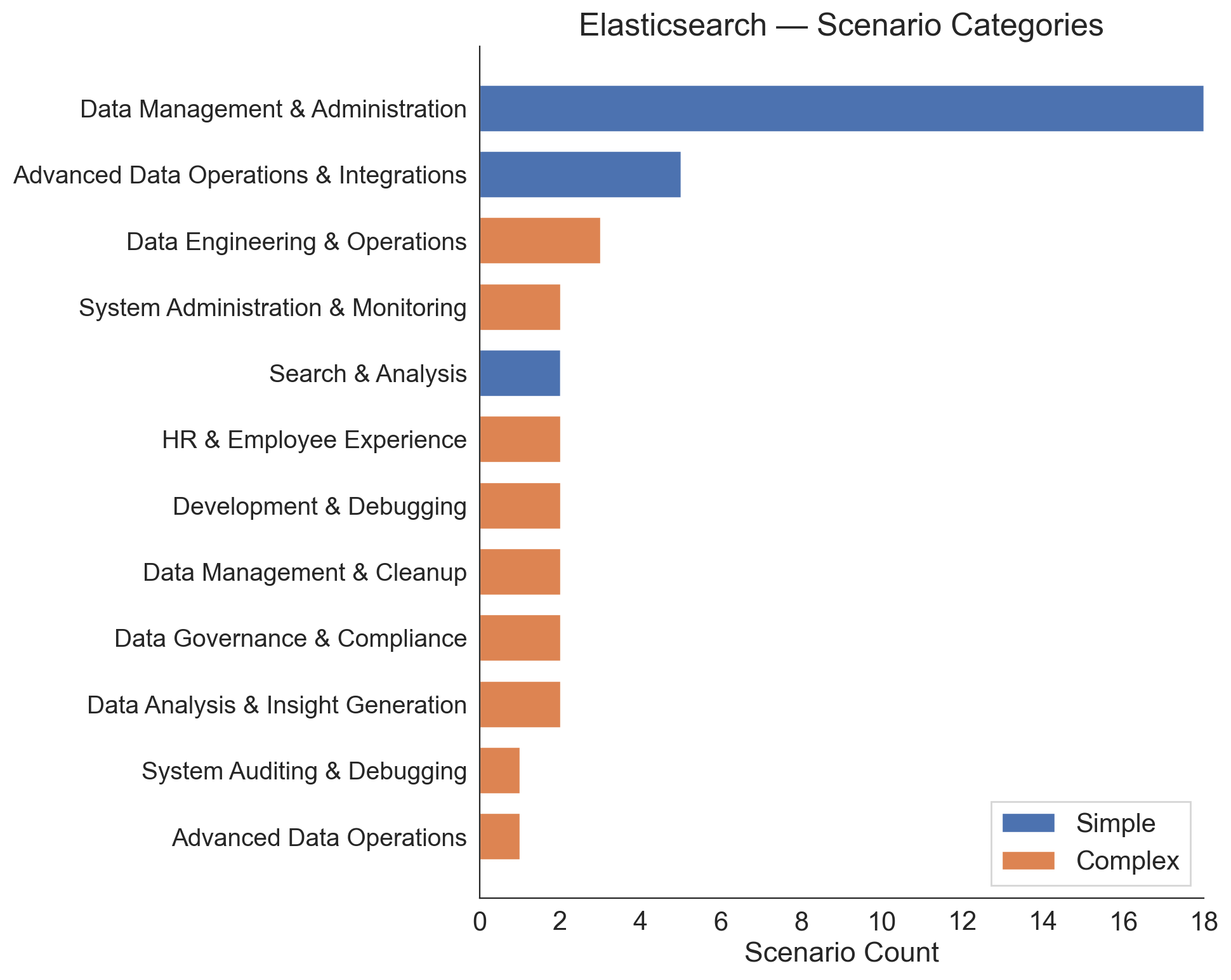}
  \caption{Scenario categories.}
\end{subfigure}
\hfill
\begin{subfigure}[t]{0.46\textwidth}
  \centering
  \includegraphics[width=\textwidth]{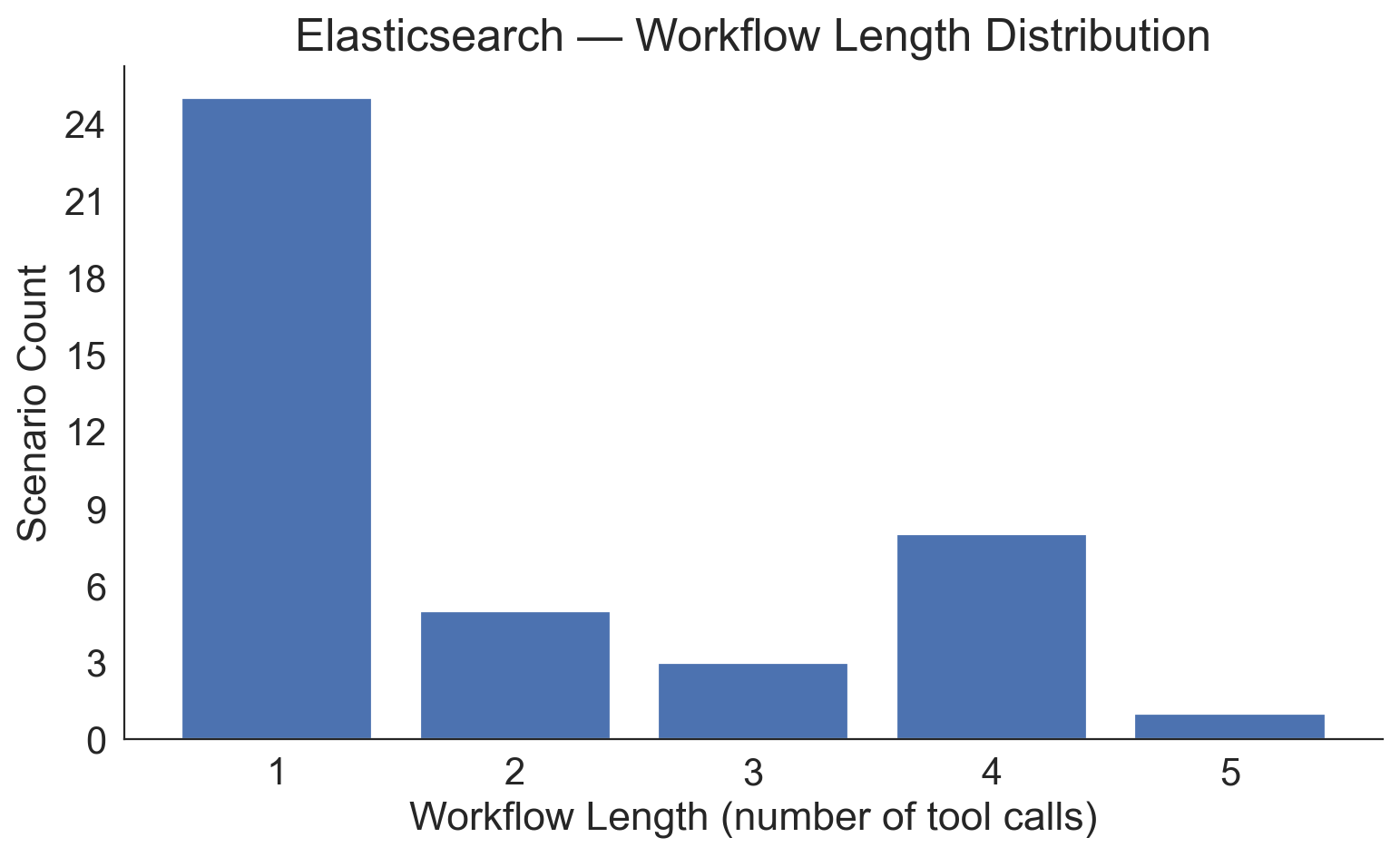}
  \caption{Workflow length distribution.}
\end{subfigure}
\\[0.5em]
\begin{subfigure}[t]{0.46\textwidth}
  \centering
  \includegraphics[width=\textwidth]{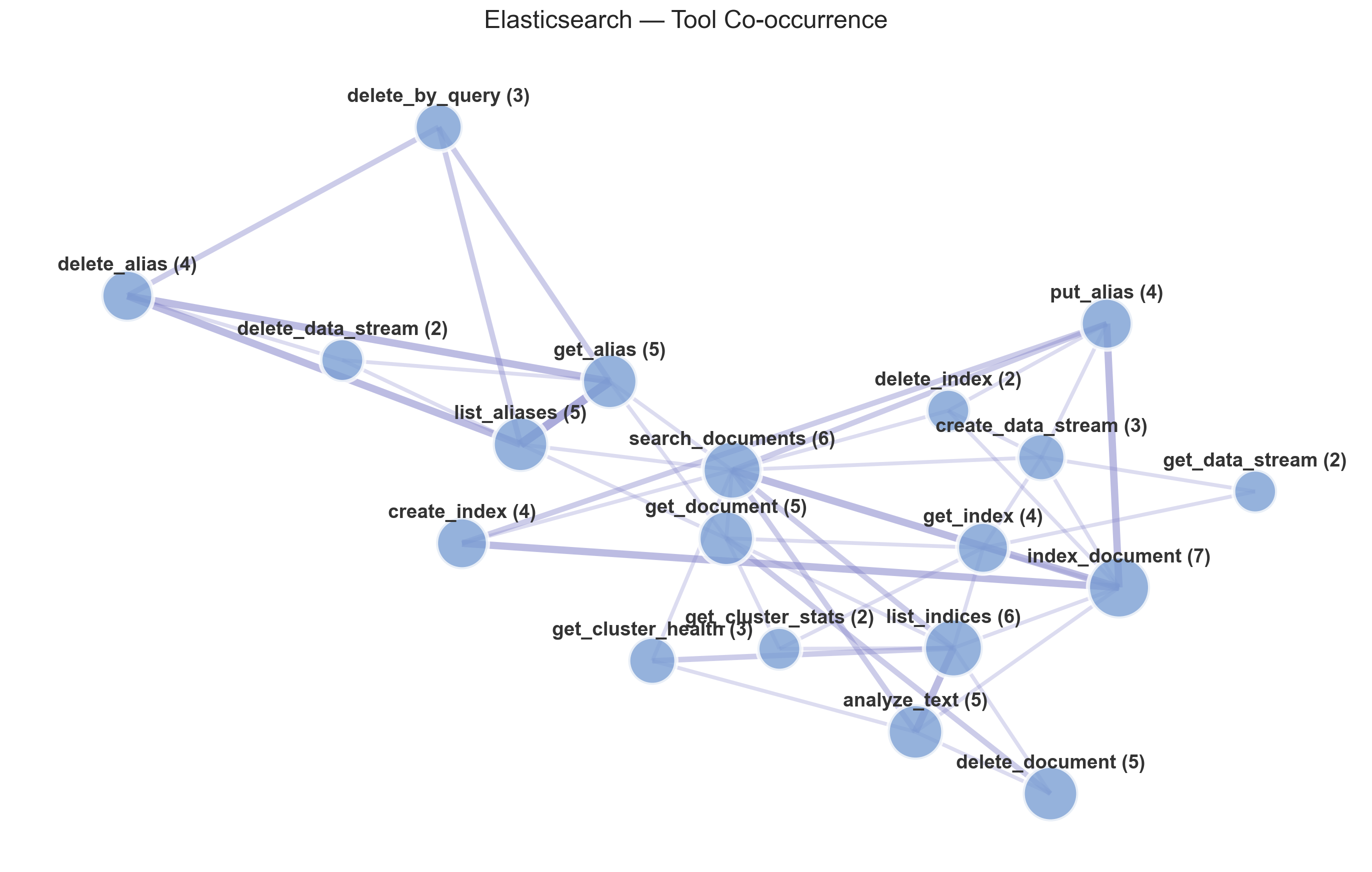}
  \caption{Tool co-occurrence graph.}
\end{subfigure}
\hfill
\begin{subfigure}[t]{0.46\textwidth}
  \centering
  \includegraphics[width=\textwidth]{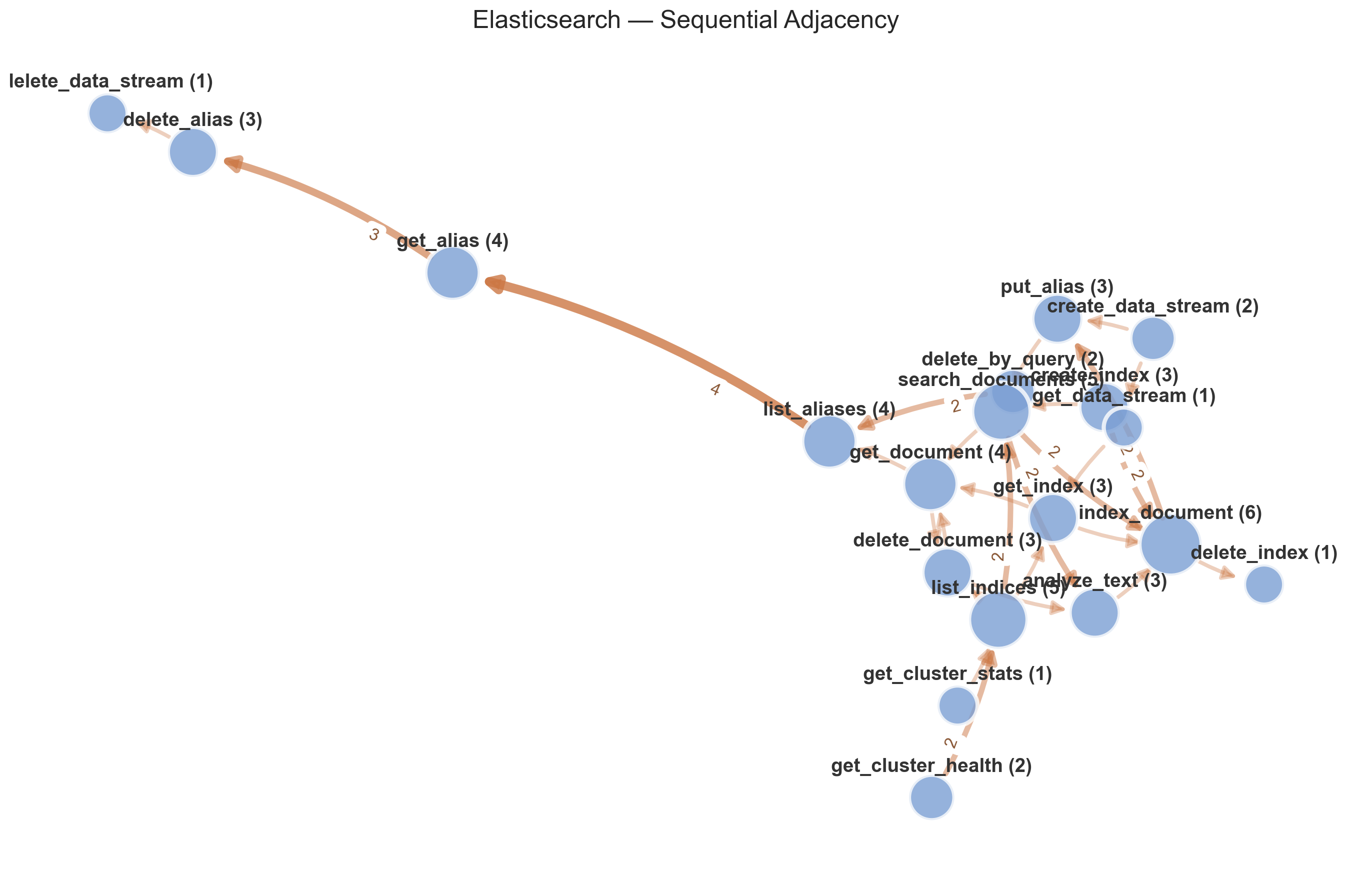}
  \caption{Sequential adjacency graph.}
\end{subfigure}
\caption{Elasticsearch (20 tools): scenario categories, workflow
lengths (mean 1.9 calls), tool co-occurrence, and sequential adjacency.}
\label{fig:elasticsearch}
\end{figure*}

\begin{figure*}[t]
\centering
\begin{subfigure}[t]{0.46\textwidth}
  \centering
  \includegraphics[width=\textwidth]{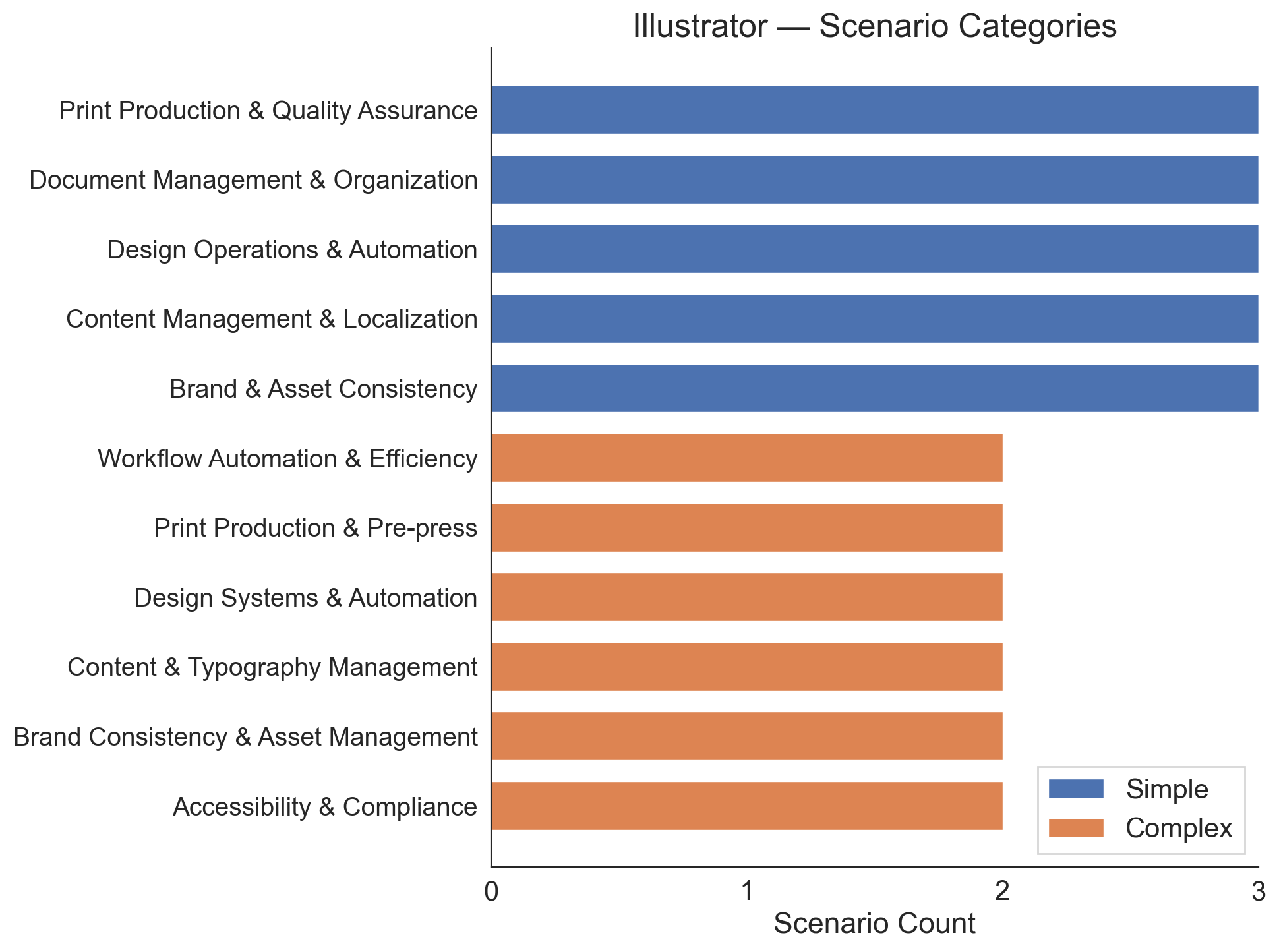}
  \caption{Scenario categories.}
\end{subfigure}
\hfill
\begin{subfigure}[t]{0.46\textwidth}
  \centering
  \includegraphics[width=\textwidth]{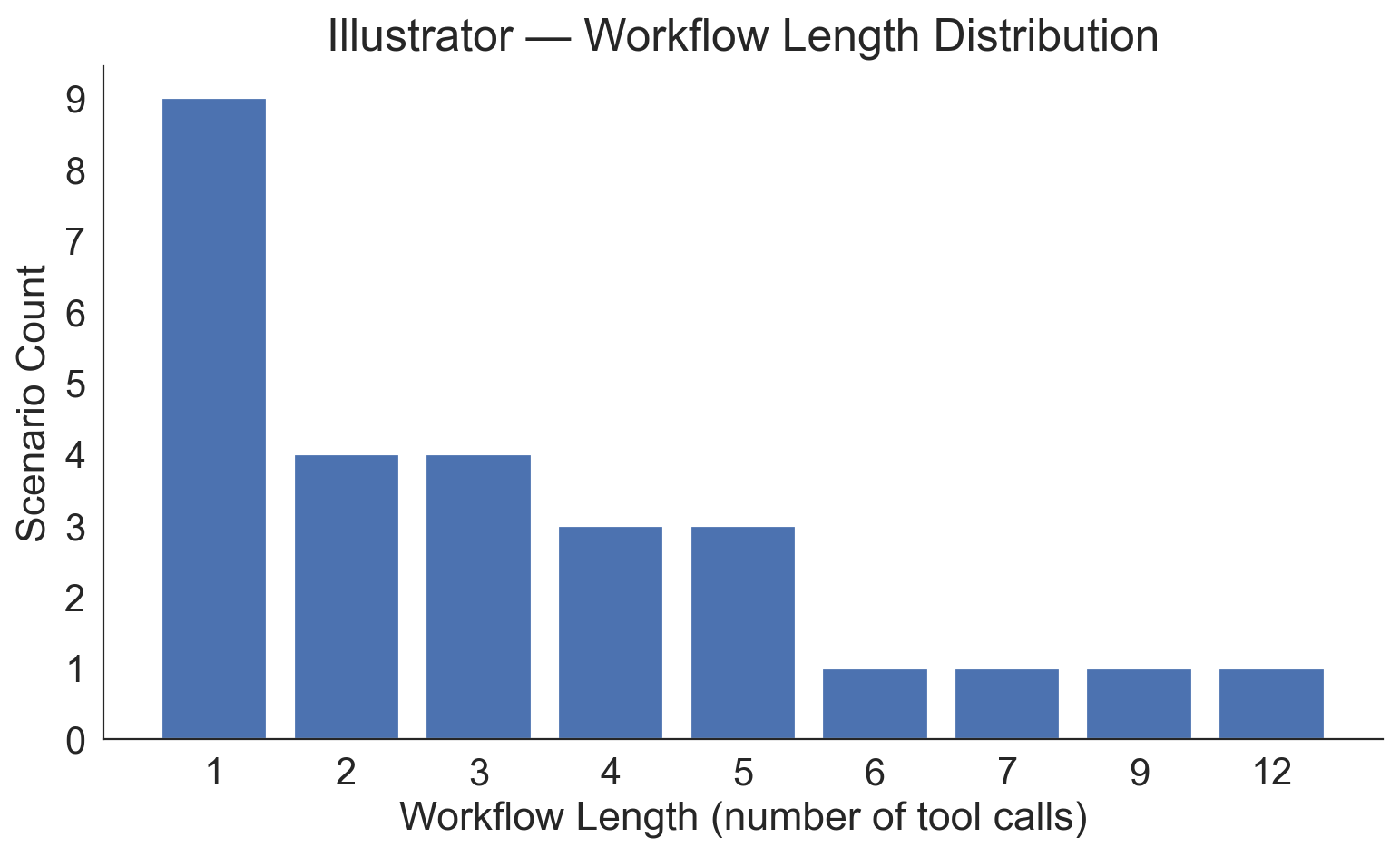}
  \caption{Workflow length distribution.}
\end{subfigure}
\\[0.5em]
\begin{subfigure}[t]{0.46\textwidth}
  \centering
  \includegraphics[width=\textwidth]{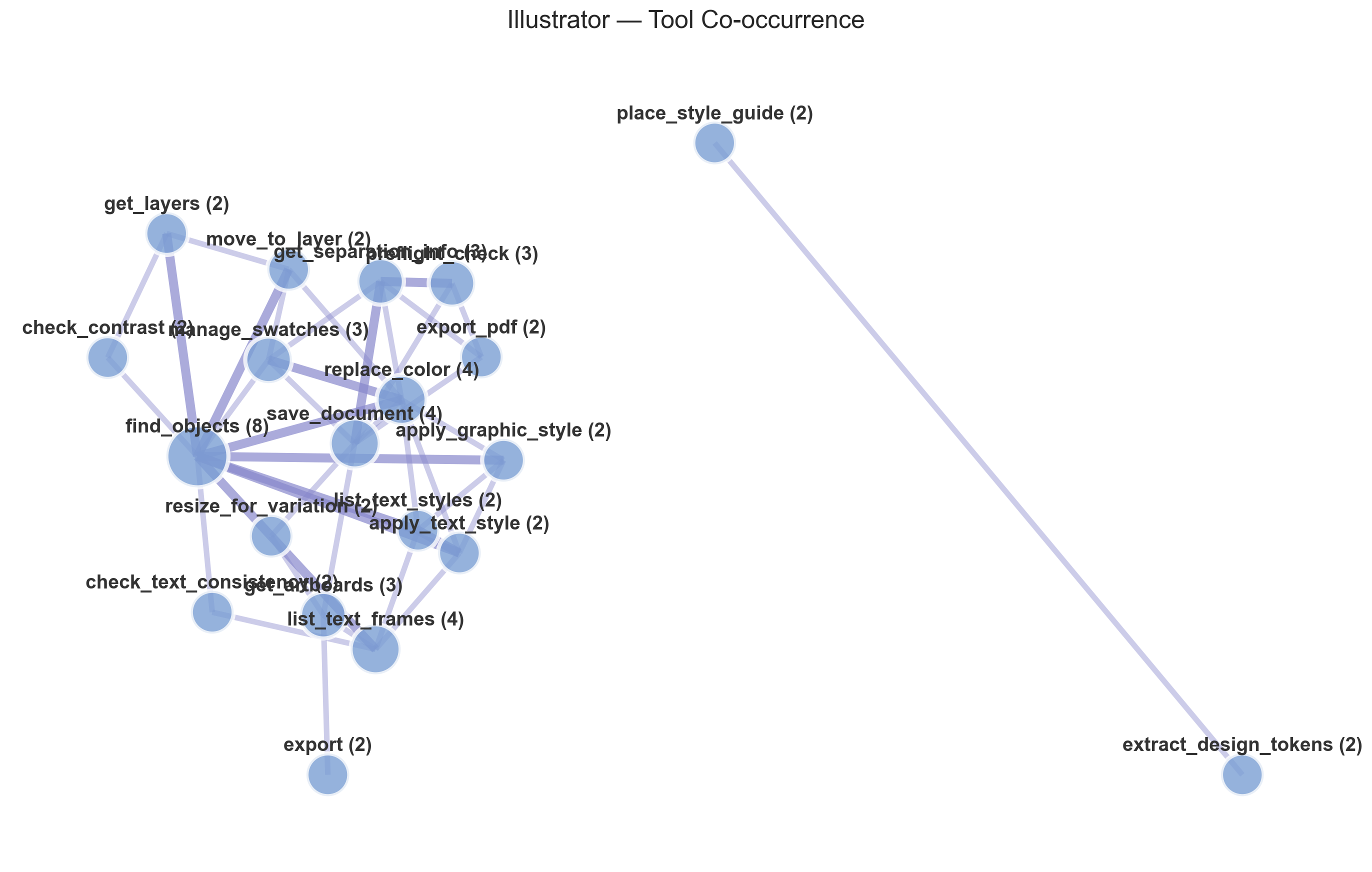}
  \caption{Tool co-occurrence graph.}
\end{subfigure}
\hfill
\begin{subfigure}[t]{0.46\textwidth}
  \centering
  \includegraphics[width=\textwidth]{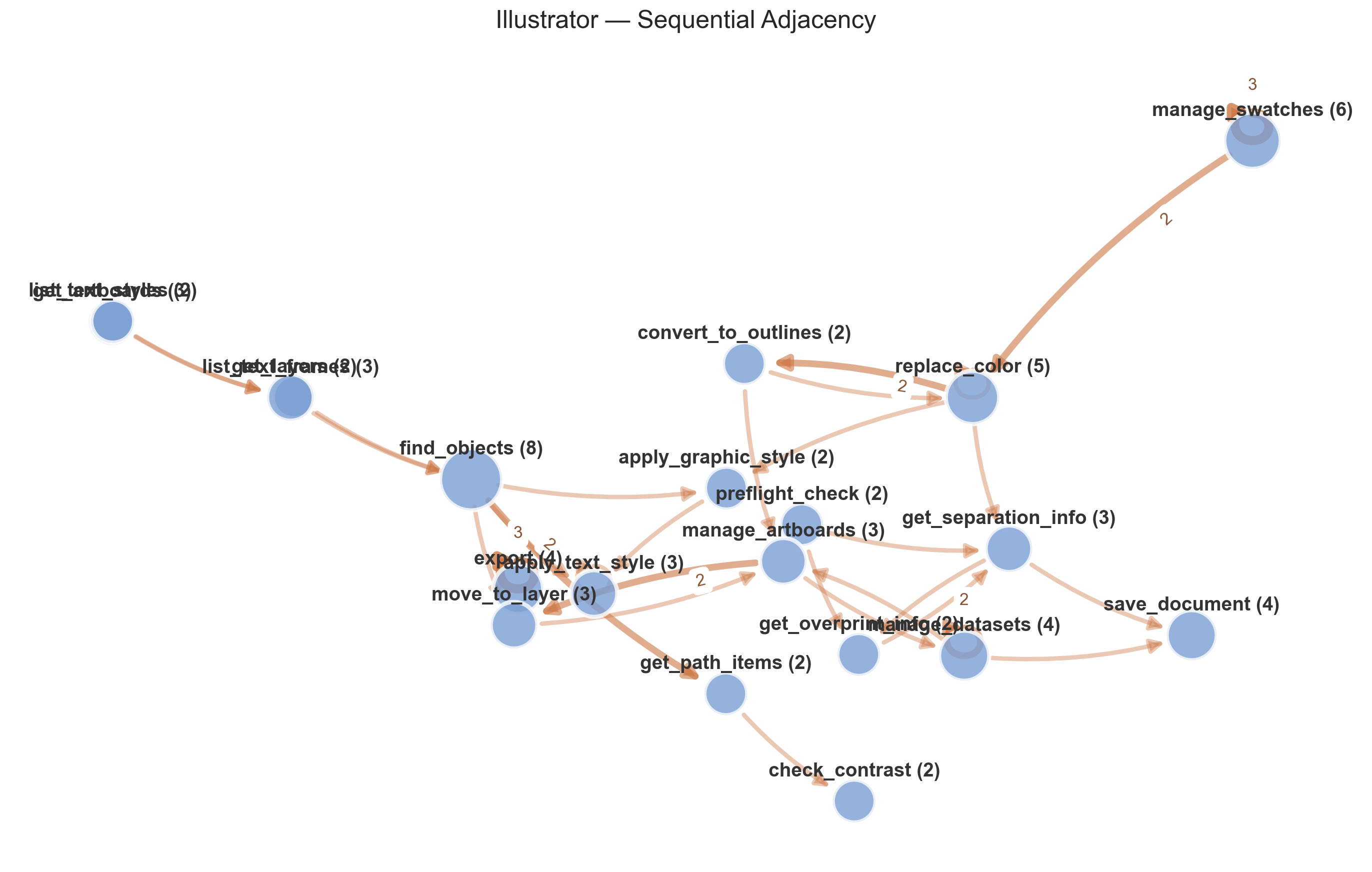}
  \caption{Sequential adjacency graph.}
\end{subfigure}
\caption{Illustrator (64 tools): scenario categories, workflow
lengths (mean 3.3 calls, max 12), tool co-occurrence, and sequential adjacency.}
\label{fig:illustrator}
\end{figure*}

\begin{figure*}[t]
\centering
\begin{subfigure}[t]{0.46\textwidth}
  \centering
  \includegraphics[width=\textwidth]{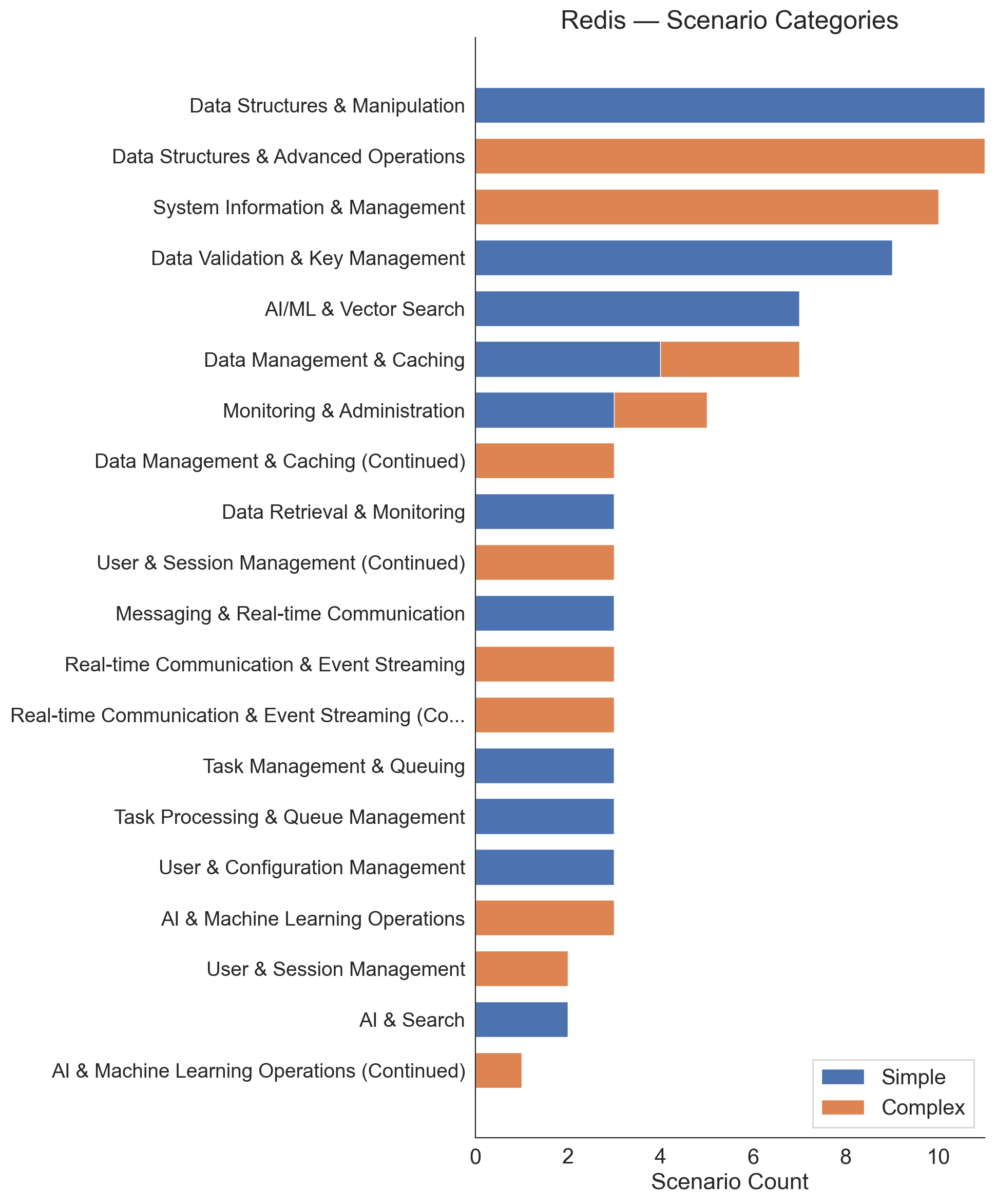}
  \caption{Scenario categories.}
\end{subfigure}
\hfill
\begin{subfigure}[t]{0.46\textwidth}
  \centering
  \includegraphics[width=\textwidth]{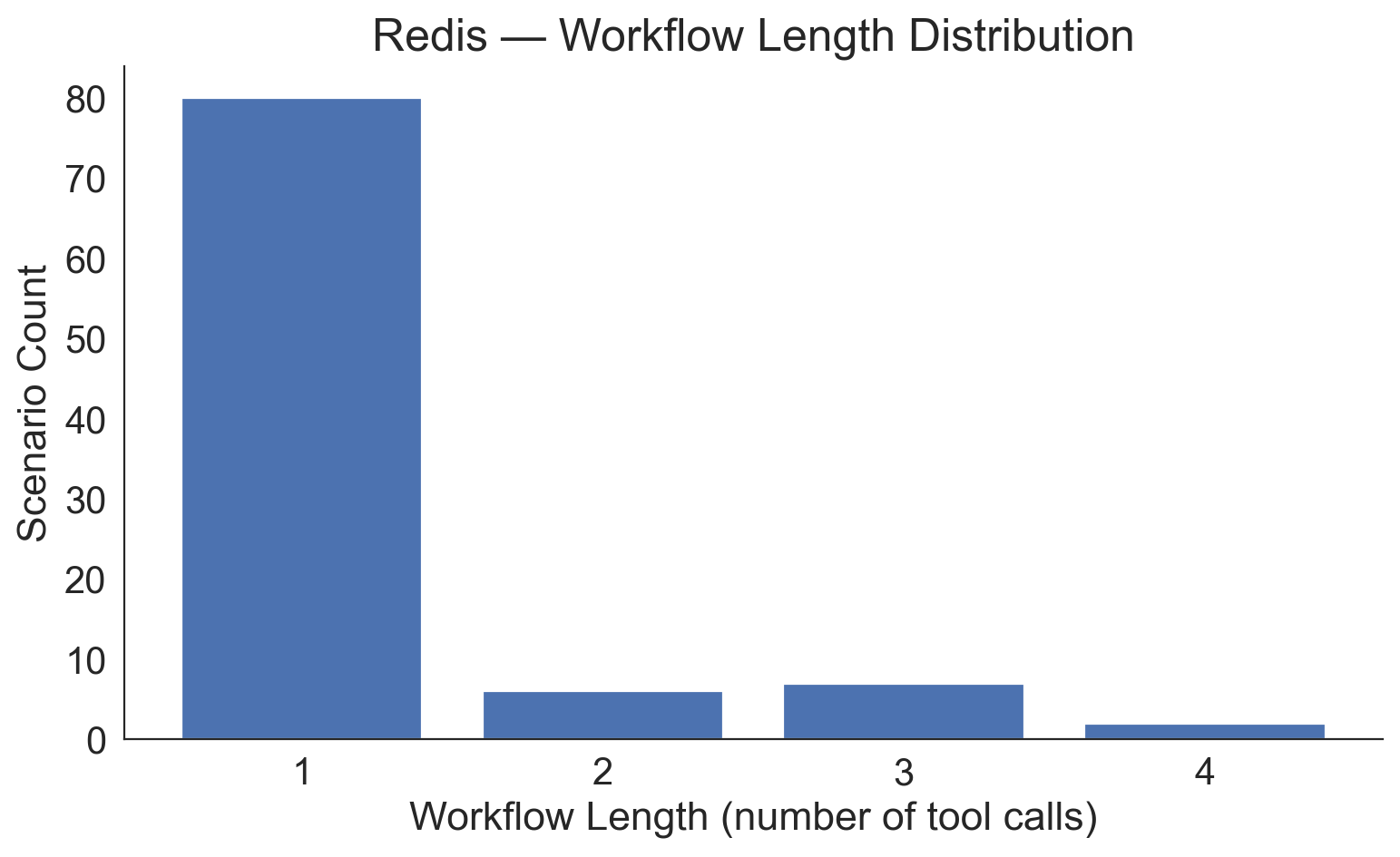}
  \caption{Workflow length distribution.}
\end{subfigure}
\\[0.5em]
\begin{subfigure}[t]{0.46\textwidth}
  \centering
  \includegraphics[width=\textwidth]{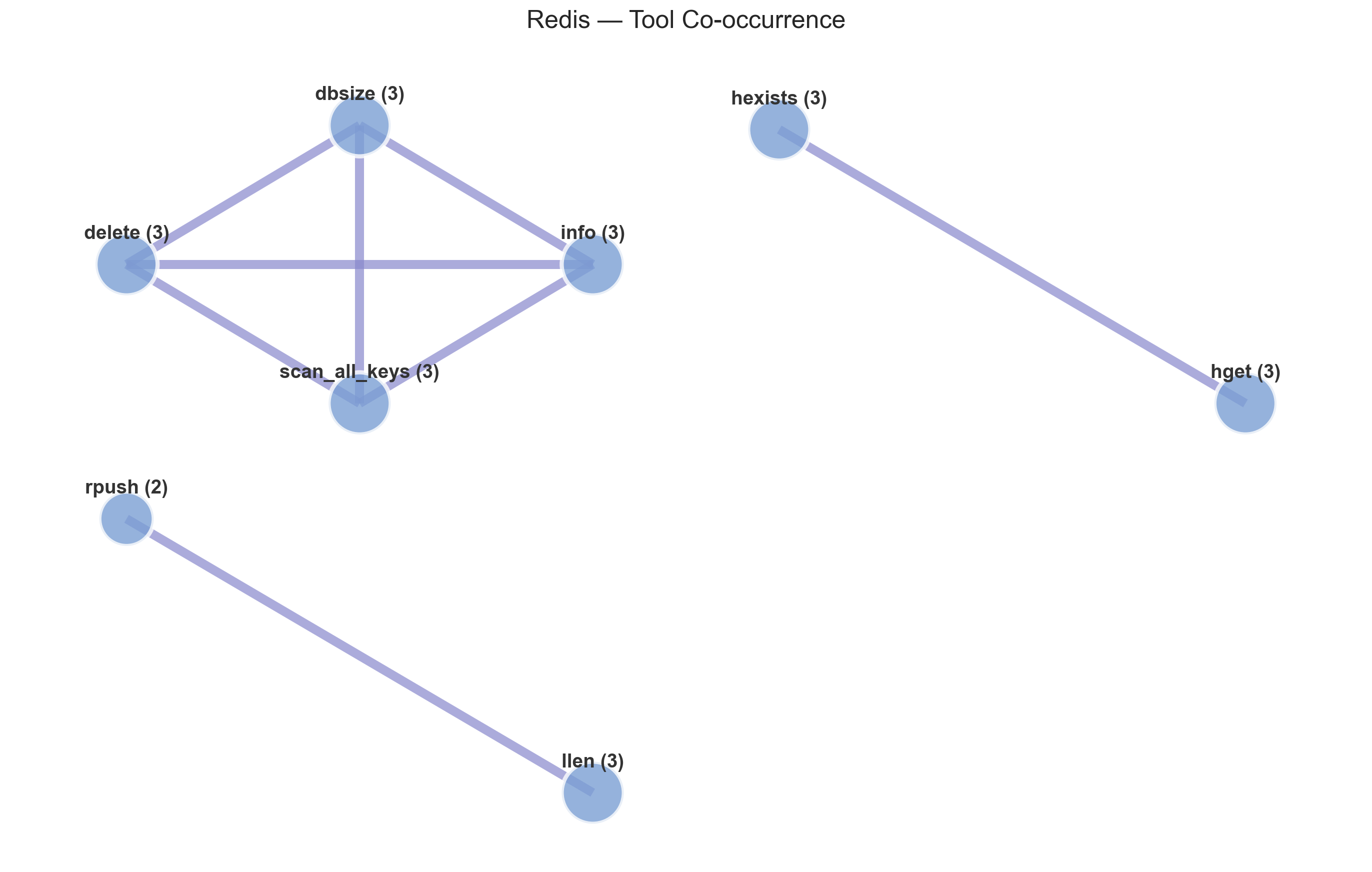}
  \caption{Tool co-occurrence graph.}
\end{subfigure}
\hfill
\begin{subfigure}[t]{0.46\textwidth}
  \centering
  \includegraphics[width=\textwidth]{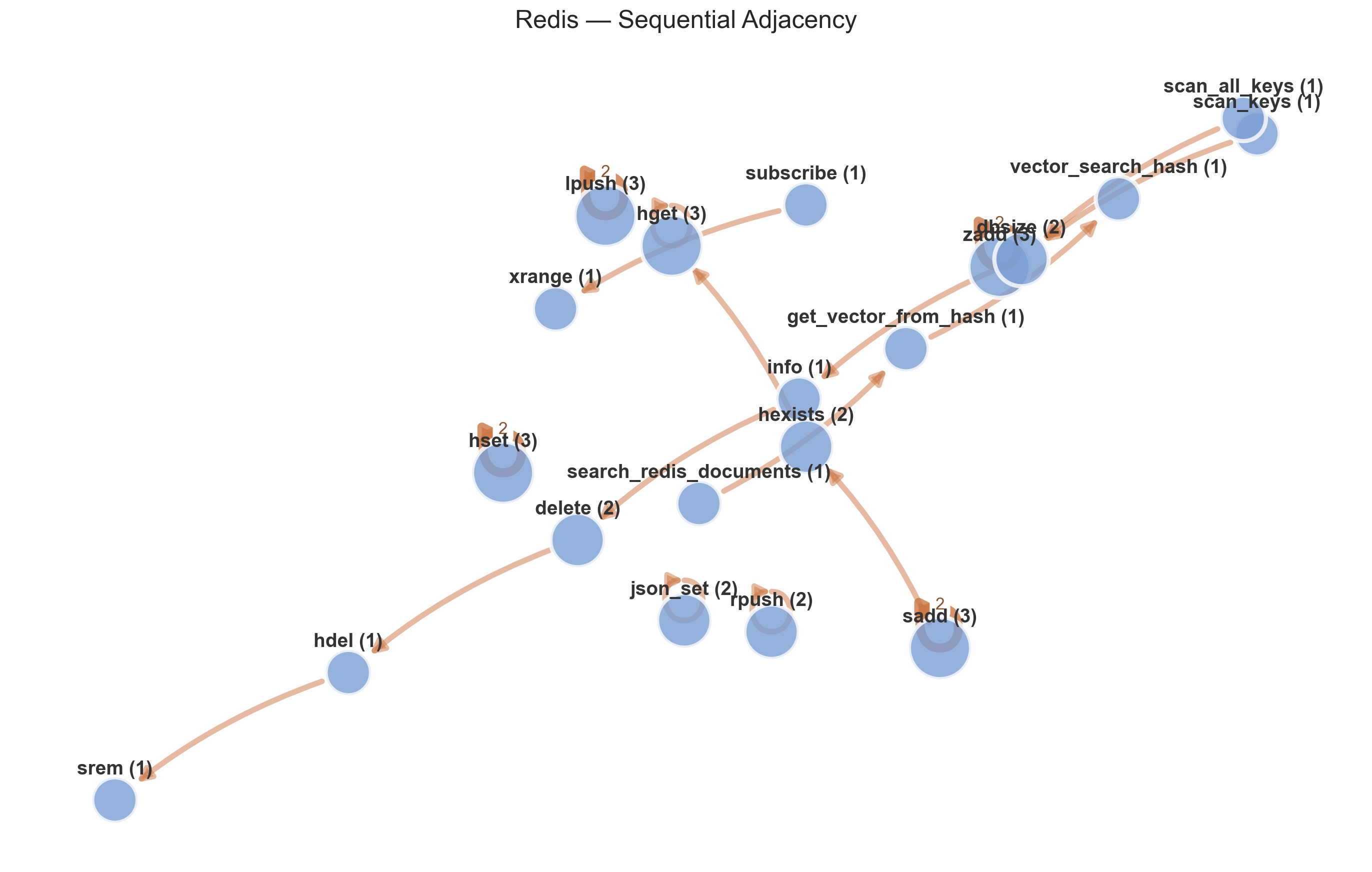}
  \caption{Sequential adjacency graph.}
\end{subfigure}
\caption{Redis (47 tools): scenario categories, workflow
lengths (mean 1.3 calls), tool co-occurrence, and sequential adjacency.}
\label{fig:redis}
\end{figure*}

\begin{figure*}[t]
\centering
\begin{subfigure}[t]{0.46\textwidth}
  \centering
  \includegraphics[width=\textwidth]{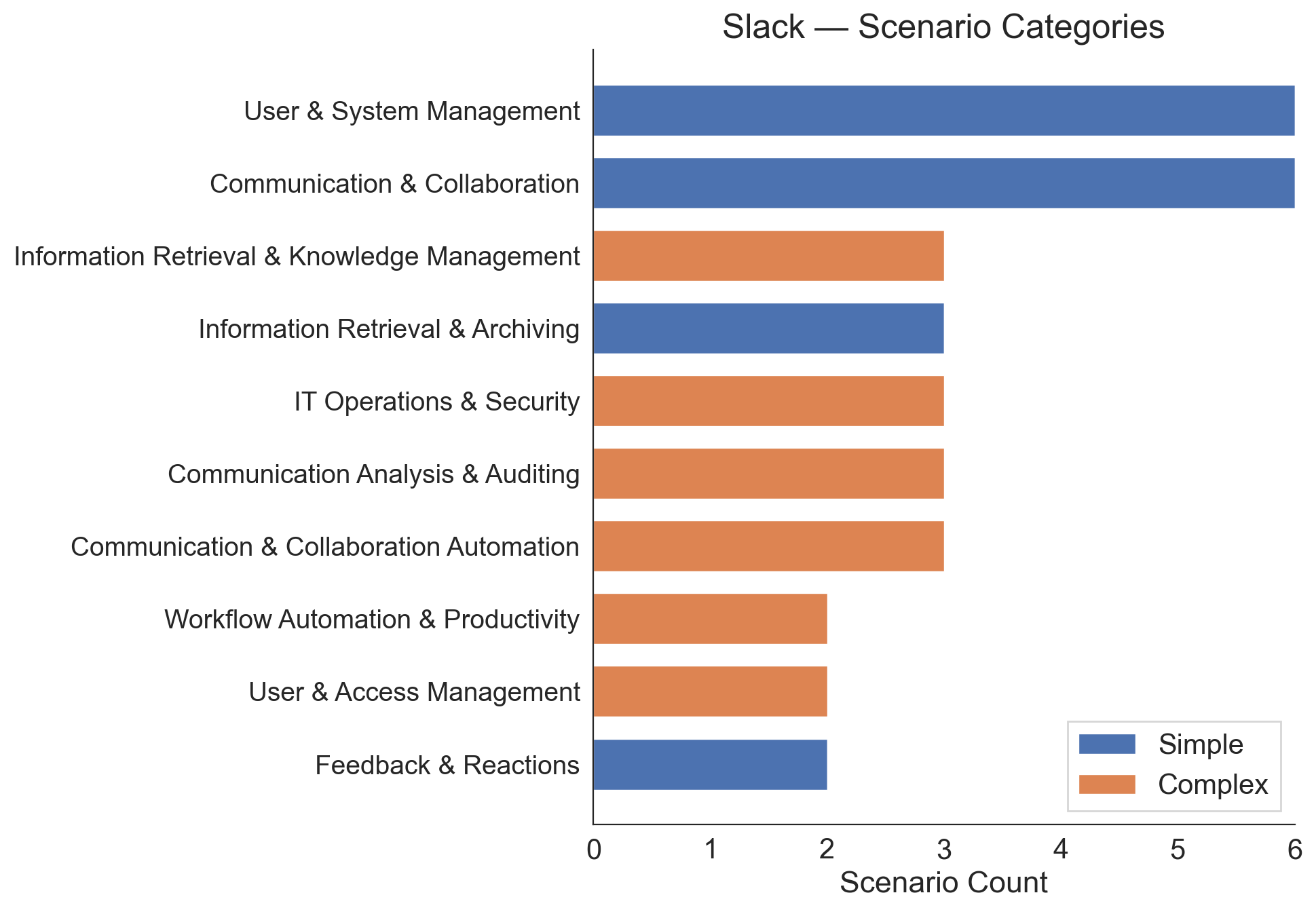}
  \caption{Scenario categories.}
\end{subfigure}
\hfill
\begin{subfigure}[t]{0.46\textwidth}
  \centering
  \includegraphics[width=\textwidth]{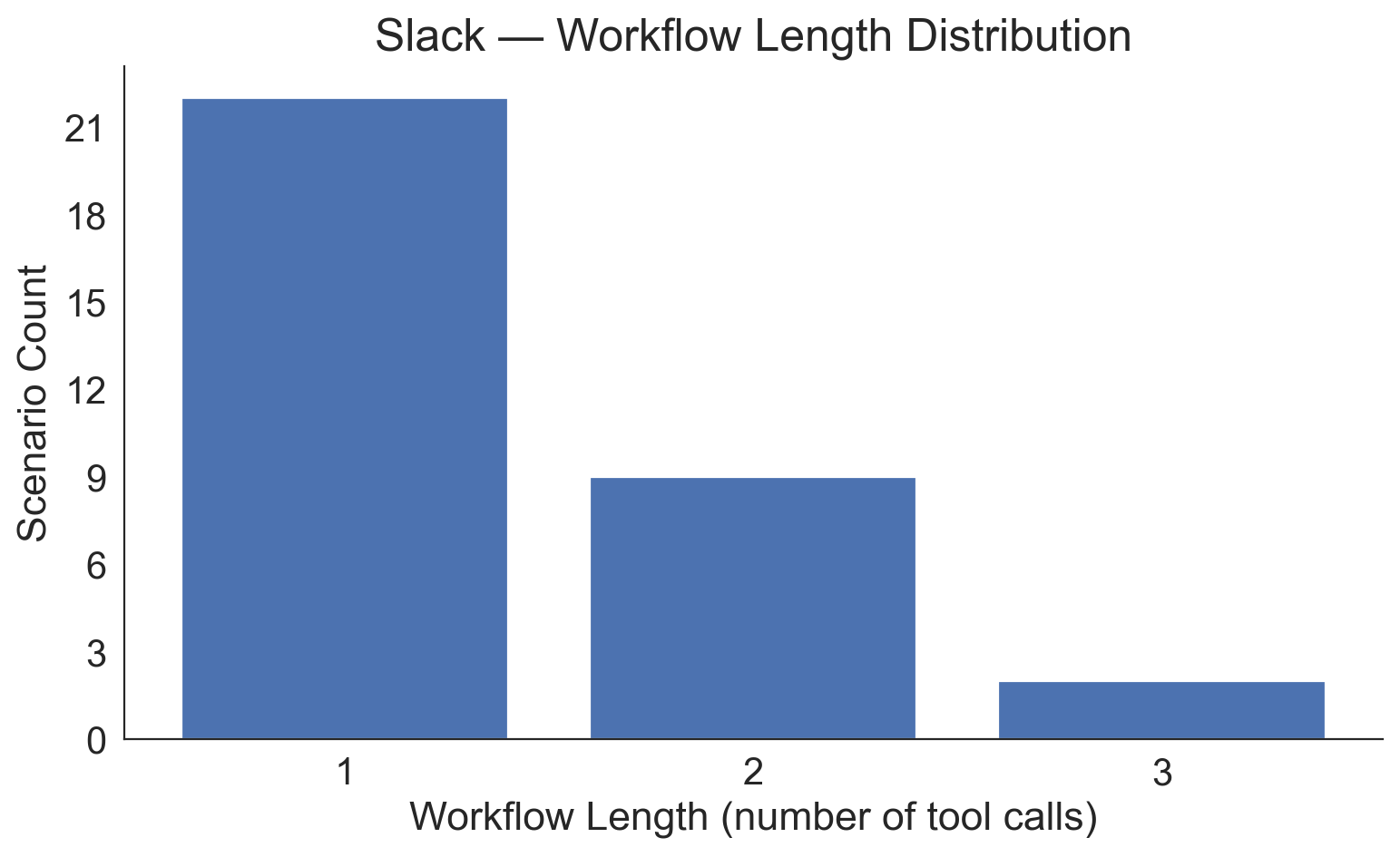}
  \caption{Workflow length distribution.}
\end{subfigure}
\\[0.5em]
\begin{subfigure}[t]{0.46\textwidth}
  \centering
  \includegraphics[width=\textwidth]{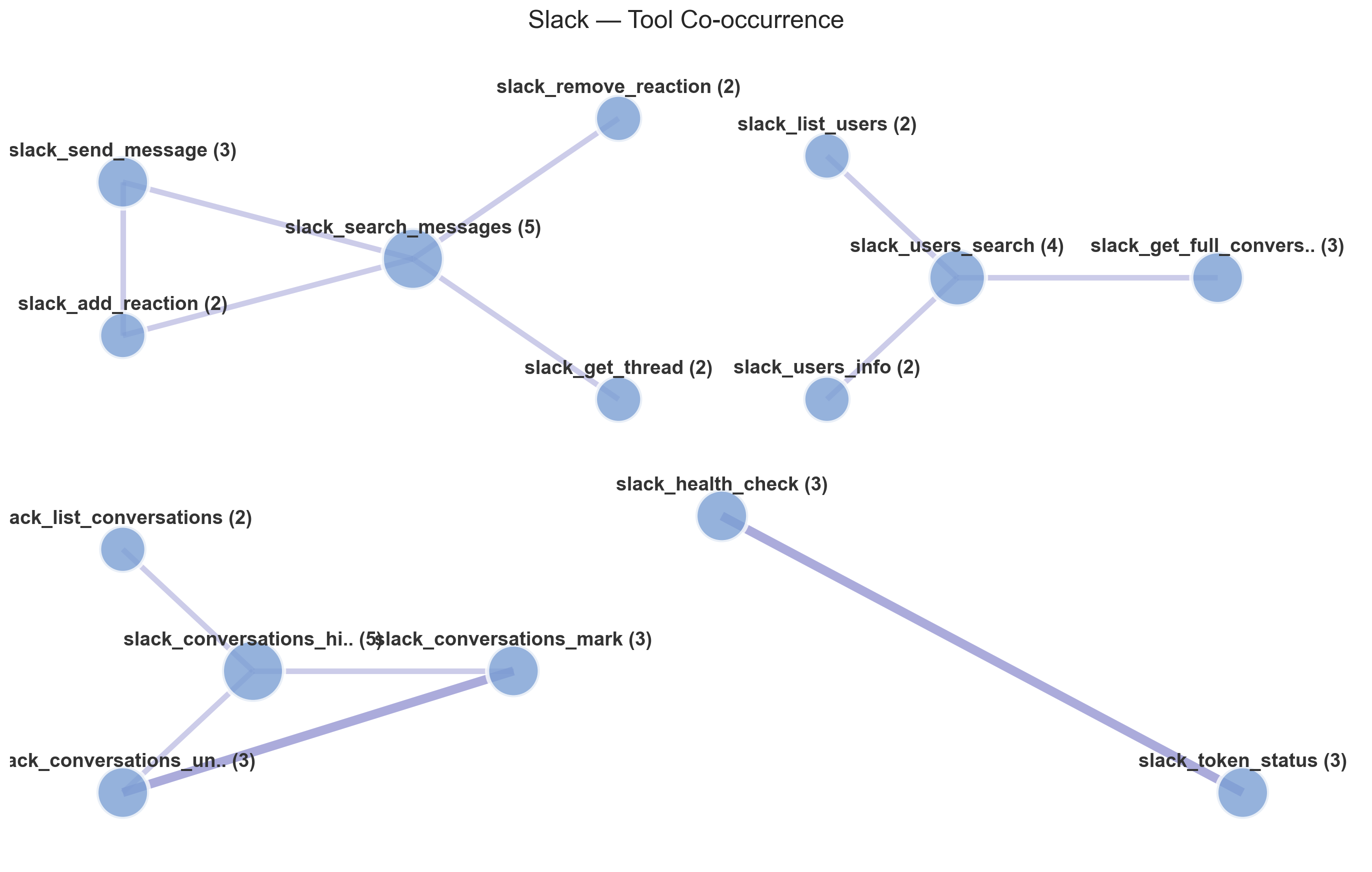}
  \caption{Tool co-occurrence graph.}
\end{subfigure}
\hfill
\begin{subfigure}[t]{0.46\textwidth}
  \centering
  \includegraphics[width=\textwidth]{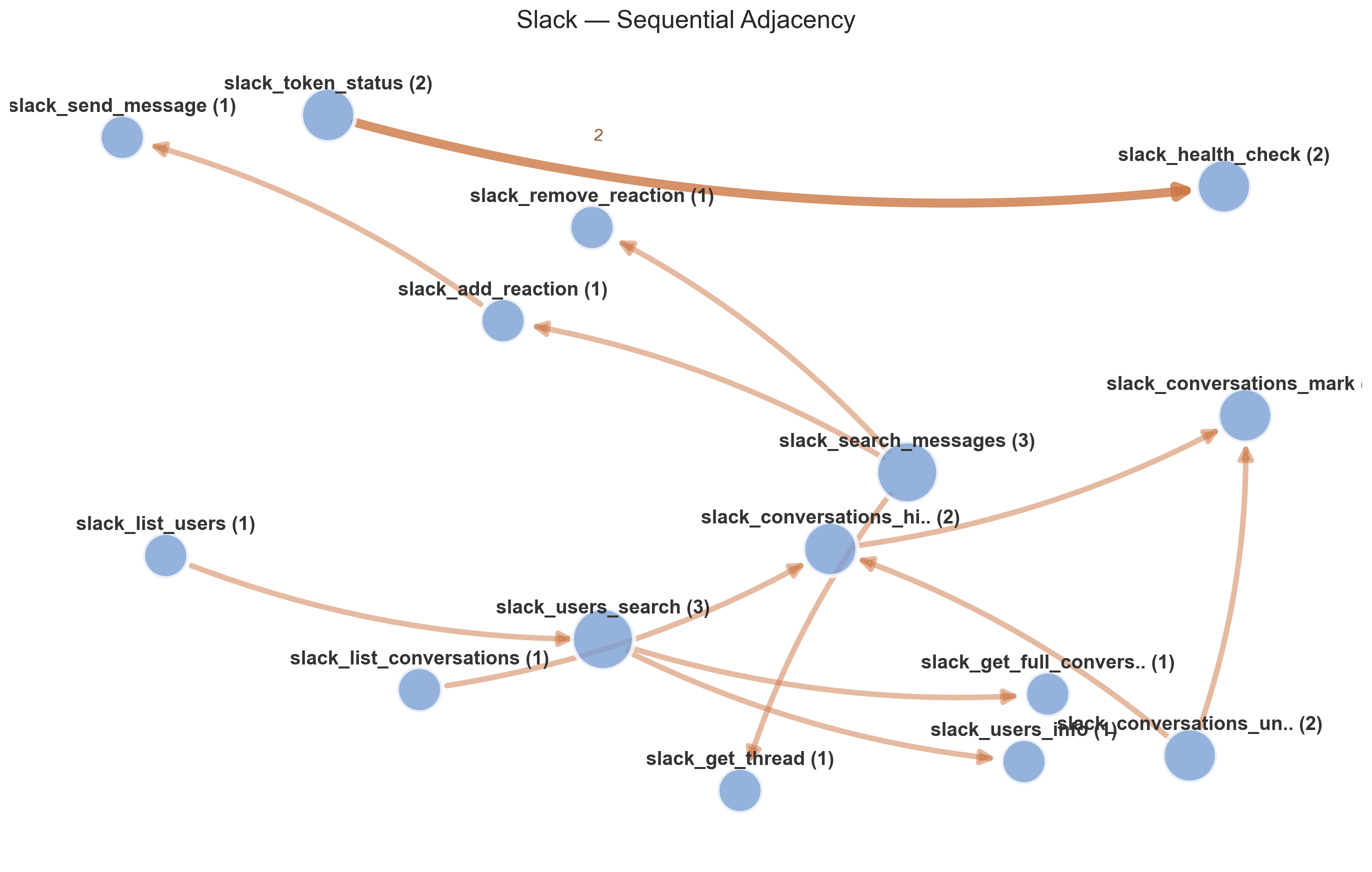}
  \caption{Sequential adjacency graph.}
\end{subfigure}
\caption{Slack (16 tools): scenario categories, workflow
lengths (mean 1.4 calls), tool co-occurrence, and sequential adjacency.}
\label{fig:slack}
\end{figure*}

\begin{figure*}[t]
\centering
\begin{subfigure}[t]{0.46\textwidth}
  \centering
  \includegraphics[width=\textwidth]{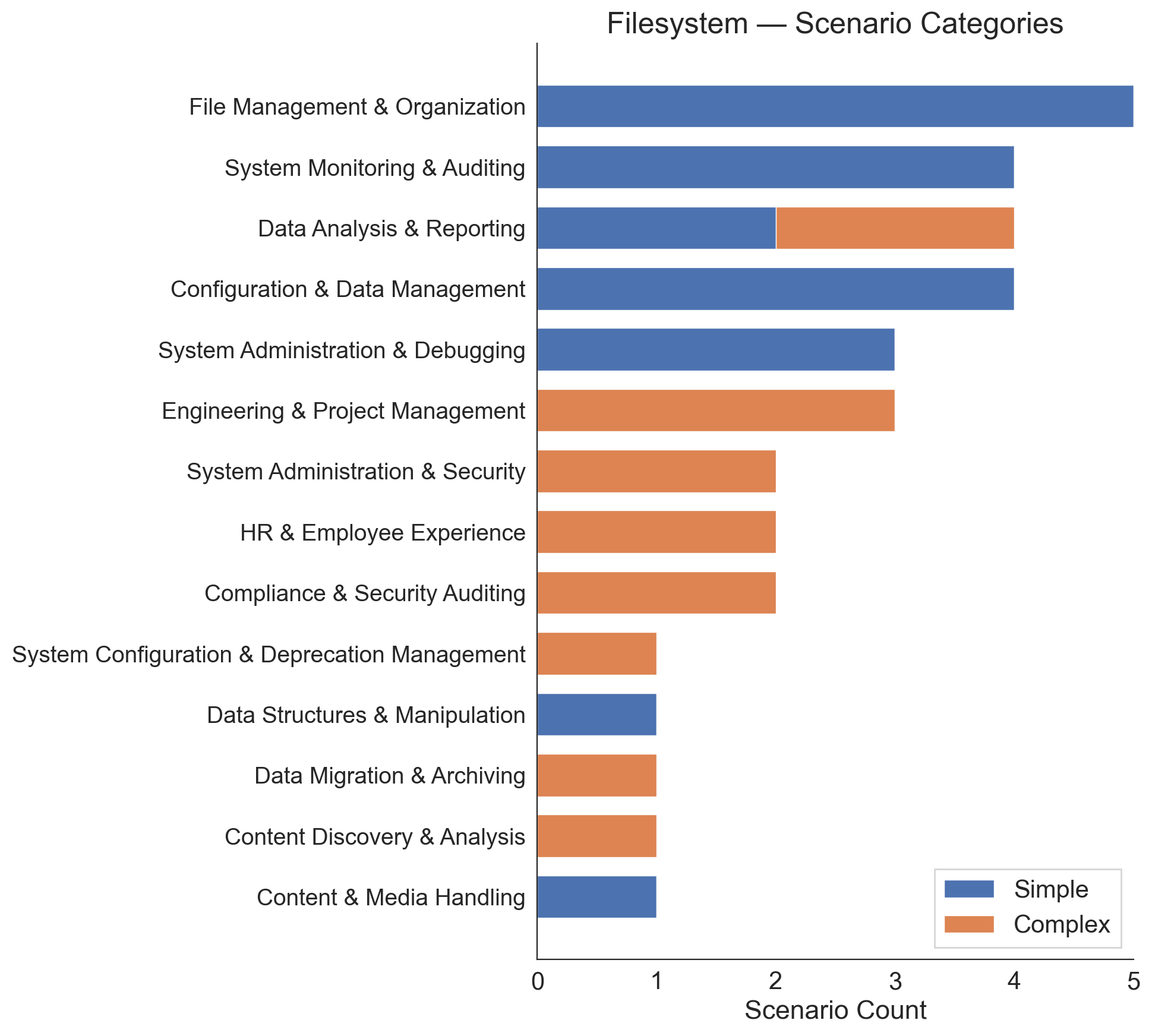}
  \caption{Scenario categories.}
\end{subfigure}
\hfill
\begin{subfigure}[t]{0.46\textwidth}
  \centering
  \includegraphics[width=\textwidth]{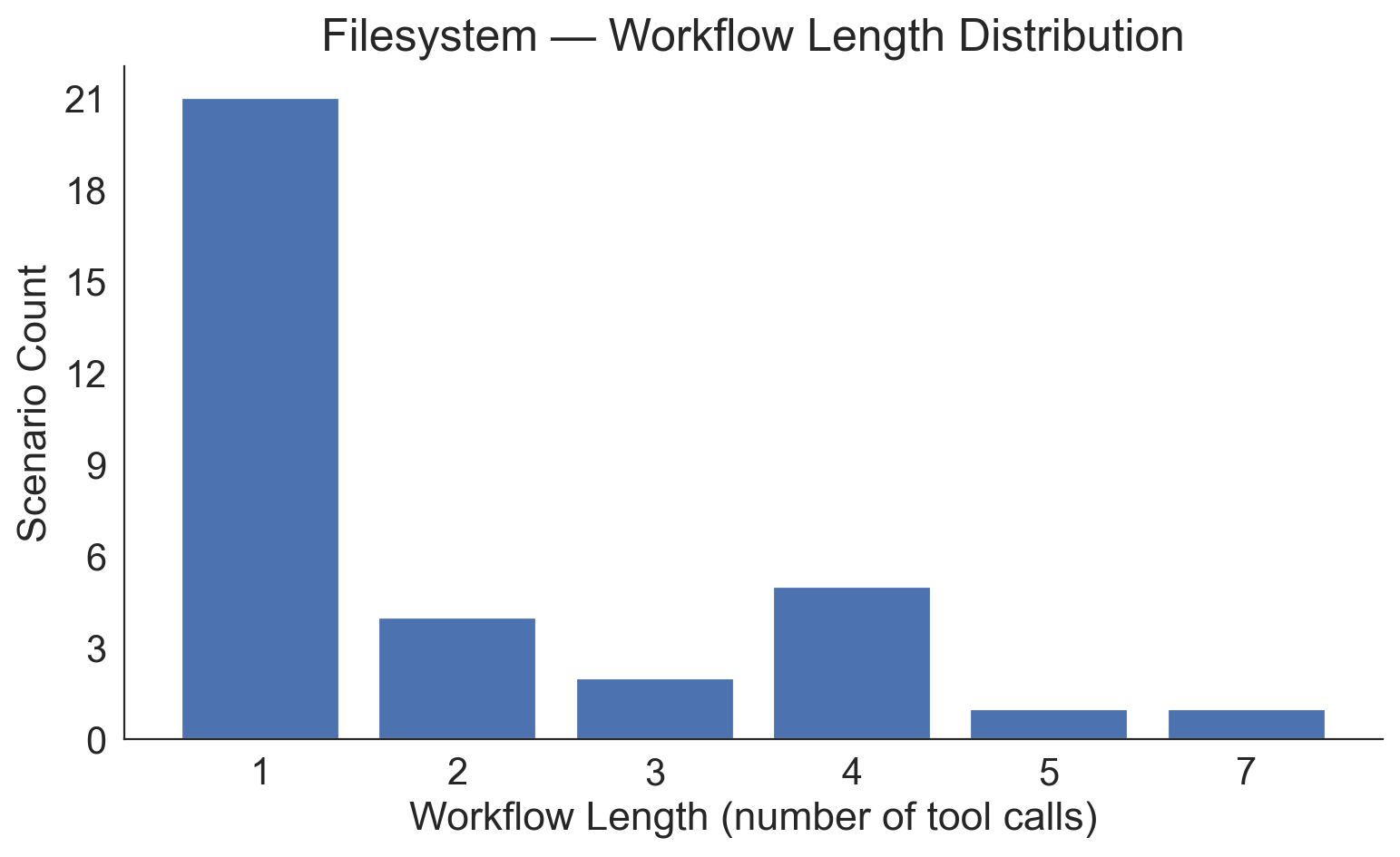}
  \caption{Workflow length distribution.}
\end{subfigure}
\\[0.5em]
\begin{subfigure}[t]{0.46\textwidth}
  \centering
  \includegraphics[width=\textwidth]{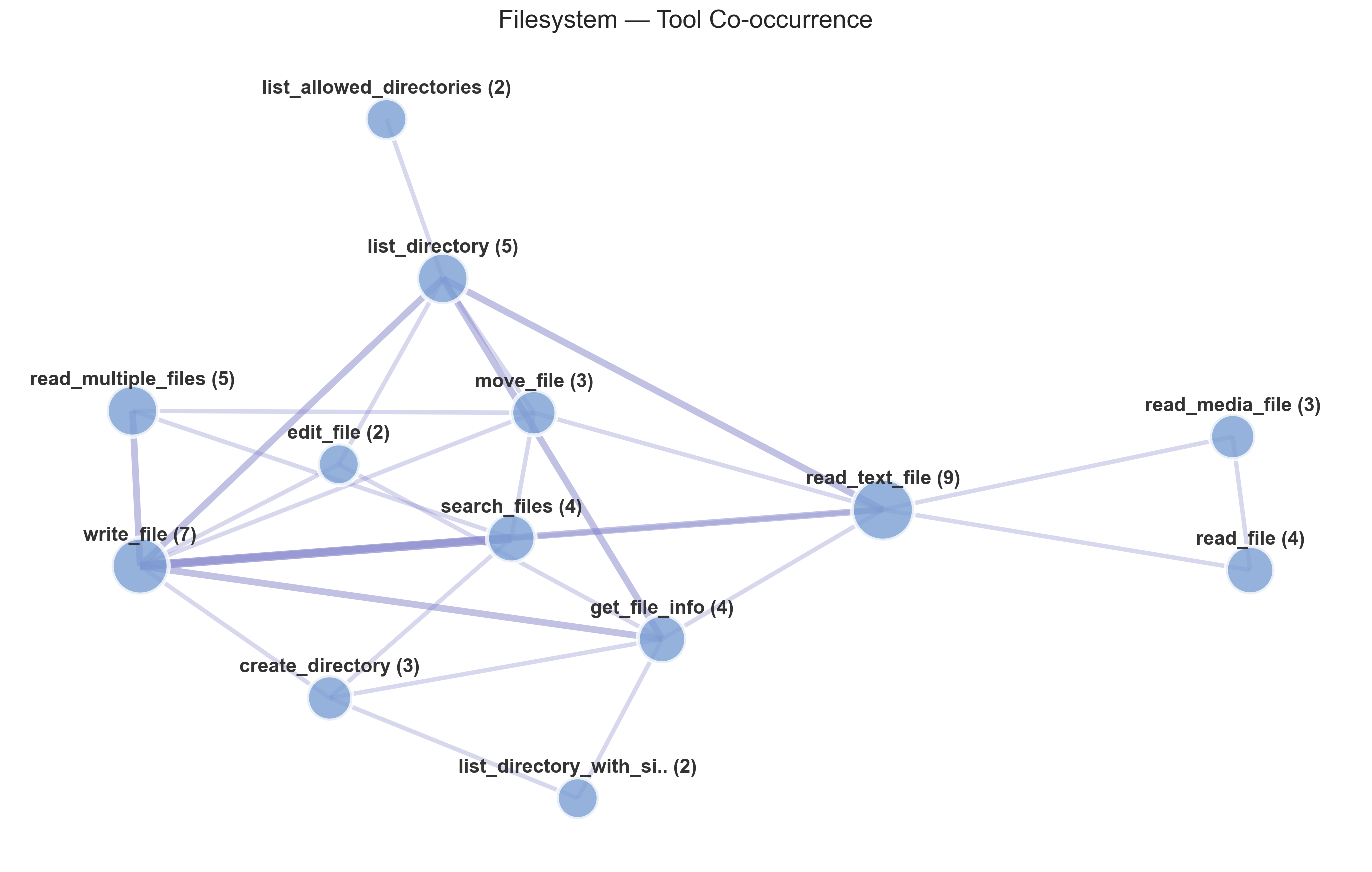}
  \caption{Tool co-occurrence graph.}
\end{subfigure}
\hfill
\begin{subfigure}[t]{0.46\textwidth}
  \centering
  \includegraphics[width=\textwidth]{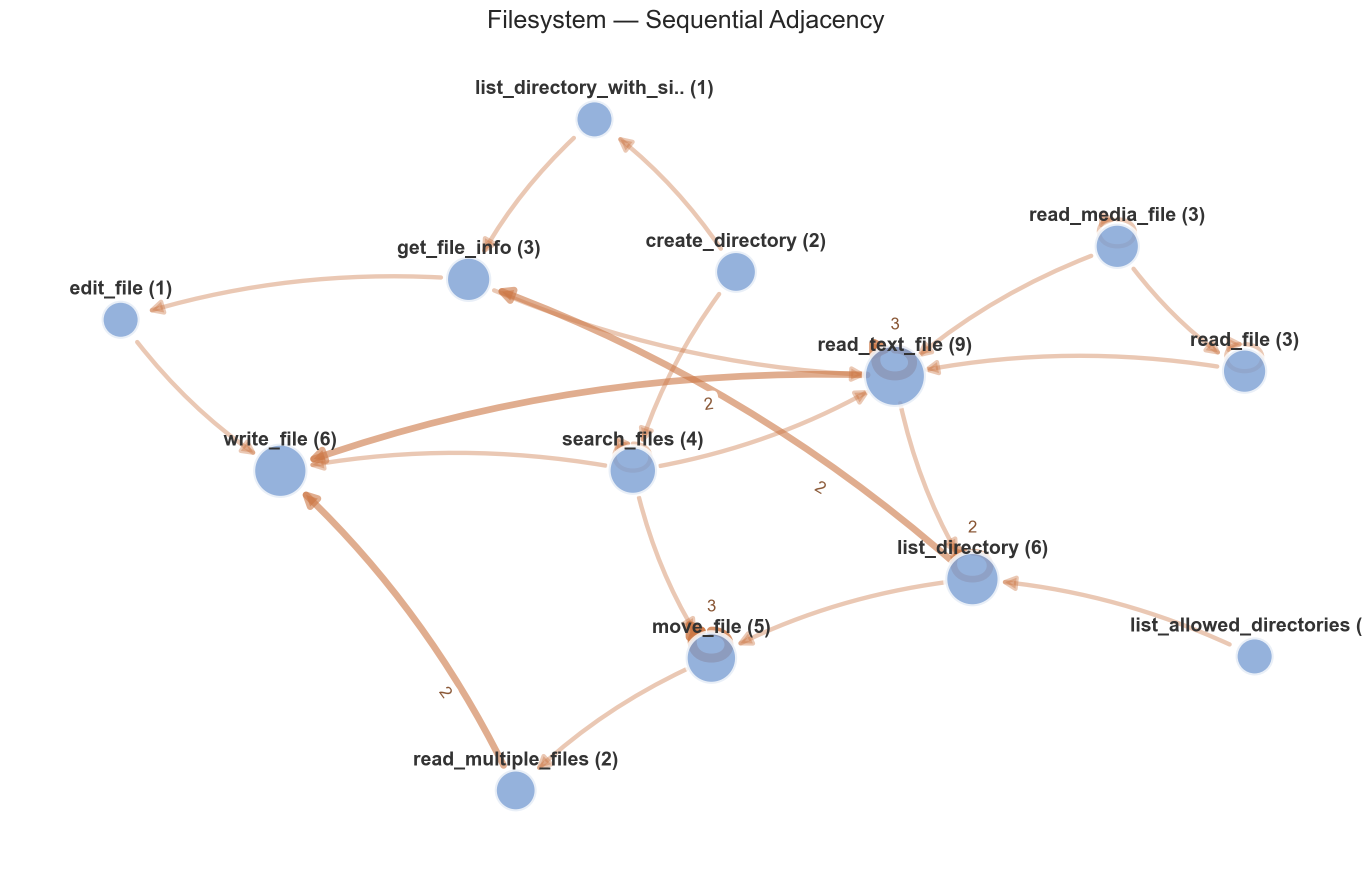}
  \caption{Sequential adjacency graph.}
\end{subfigure}
\caption{Filesystem (14 tools): scenario categories, workflow
lengths (mean 2.0 calls), tool co-occurrence, and sequential adjacency.}
\label{fig:filesystem}
\end{figure*}

\begin{figure*}[t]
\centering
\begin{subfigure}[t]{0.46\textwidth}
  \centering
  \includegraphics[width=\textwidth]{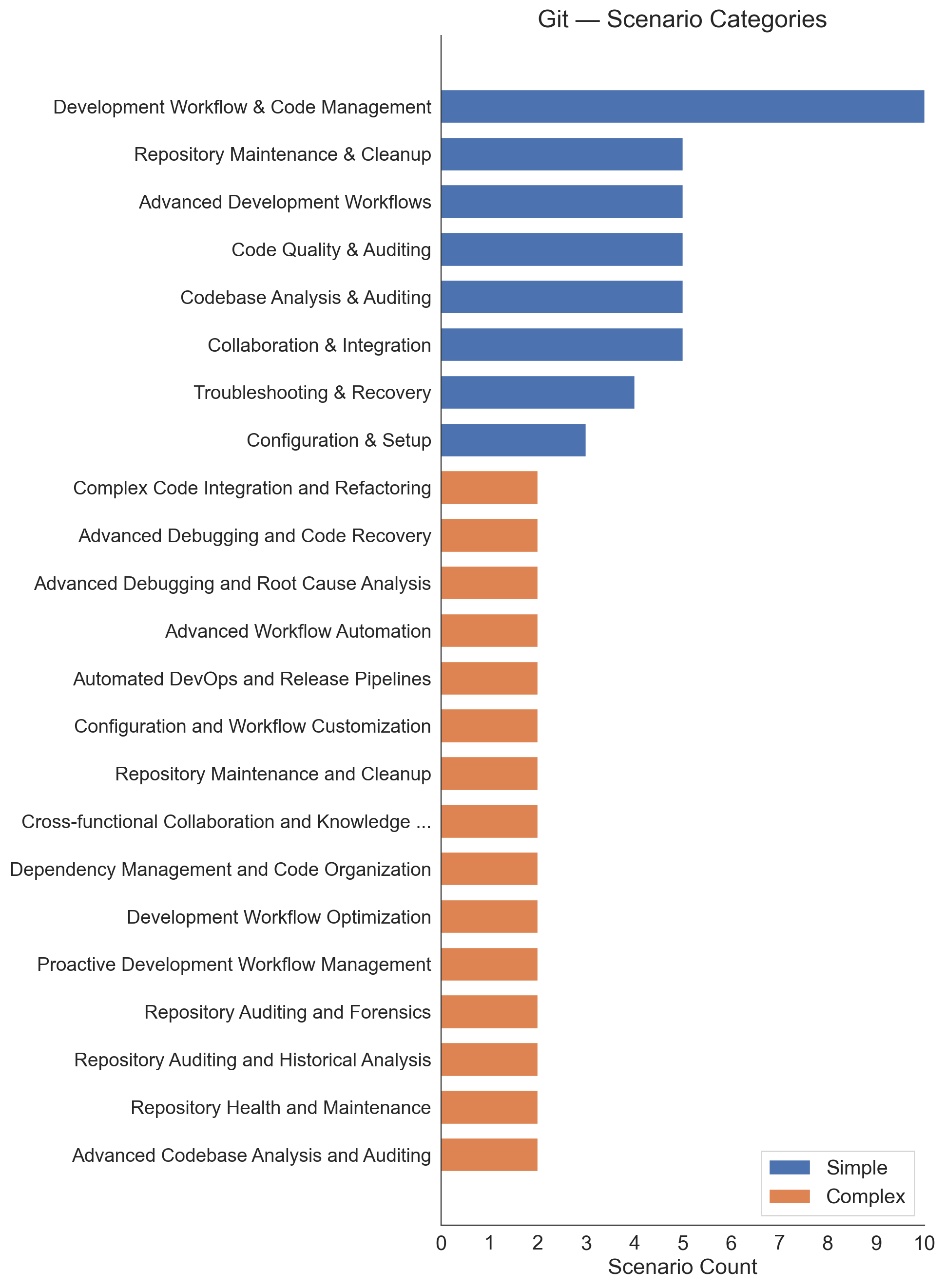}
  \caption{Scenario categories.}
\end{subfigure}
\hfill
\begin{subfigure}[t]{0.46\textwidth}
  \centering
  \includegraphics[width=\textwidth]{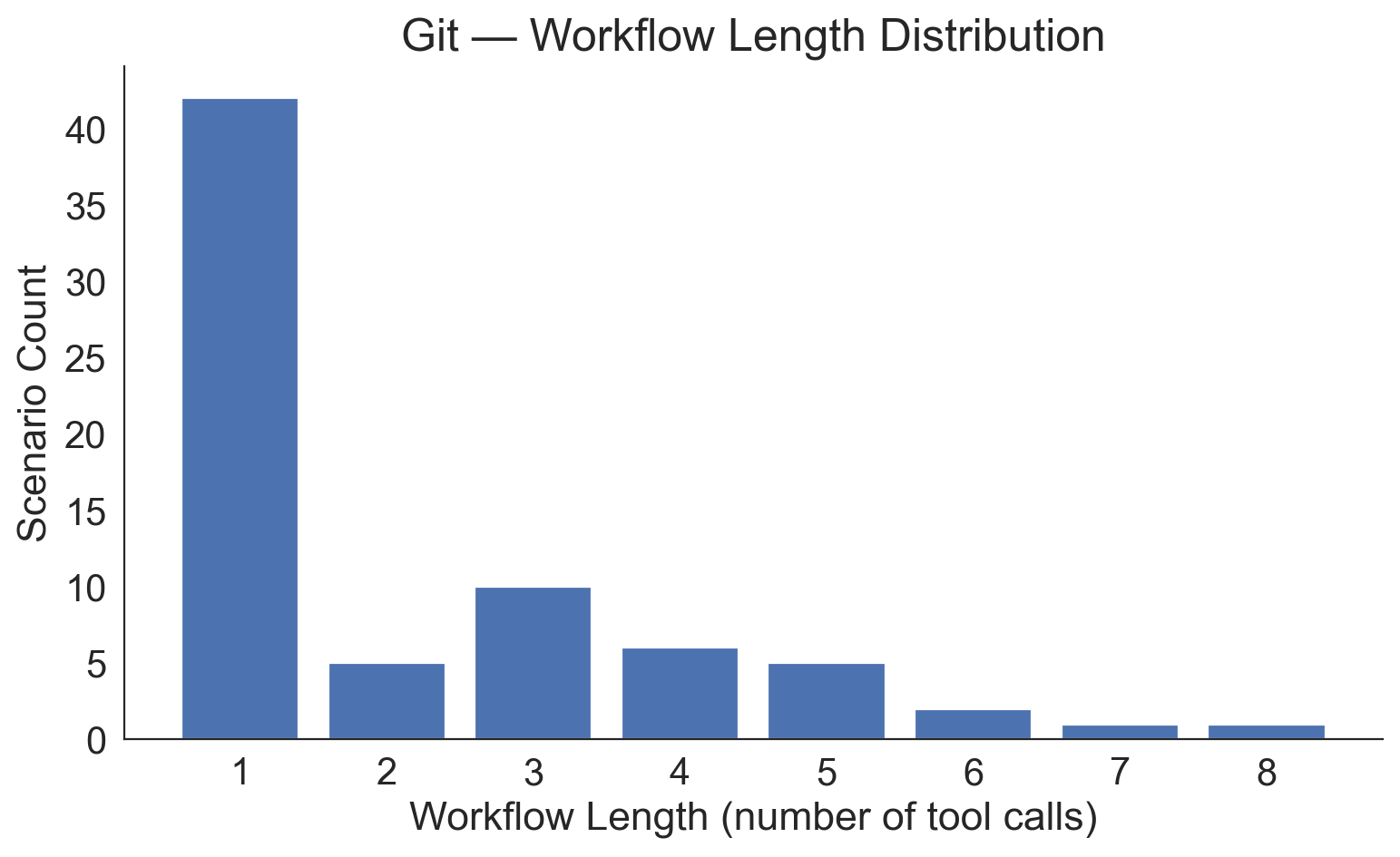}
  \caption{Workflow length distribution.}
\end{subfigure}
\\[0.5em]
\begin{subfigure}[t]{0.46\textwidth}
  \centering
  \includegraphics[width=\textwidth]{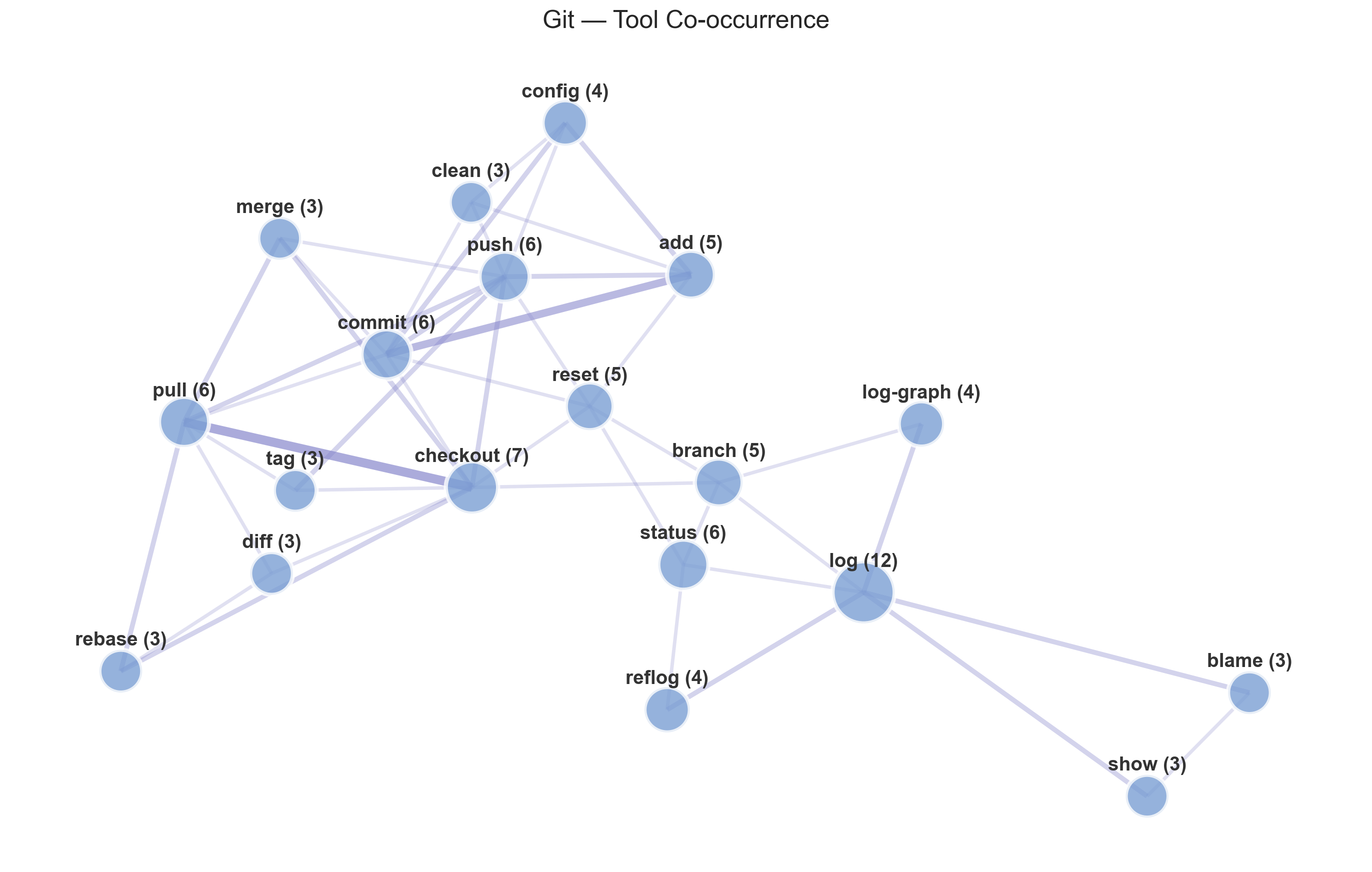}
  \caption{Tool co-occurrence graph.}
\end{subfigure}
\hfill
\begin{subfigure}[t]{0.46\textwidth}
  \centering
  \includegraphics[width=\textwidth]{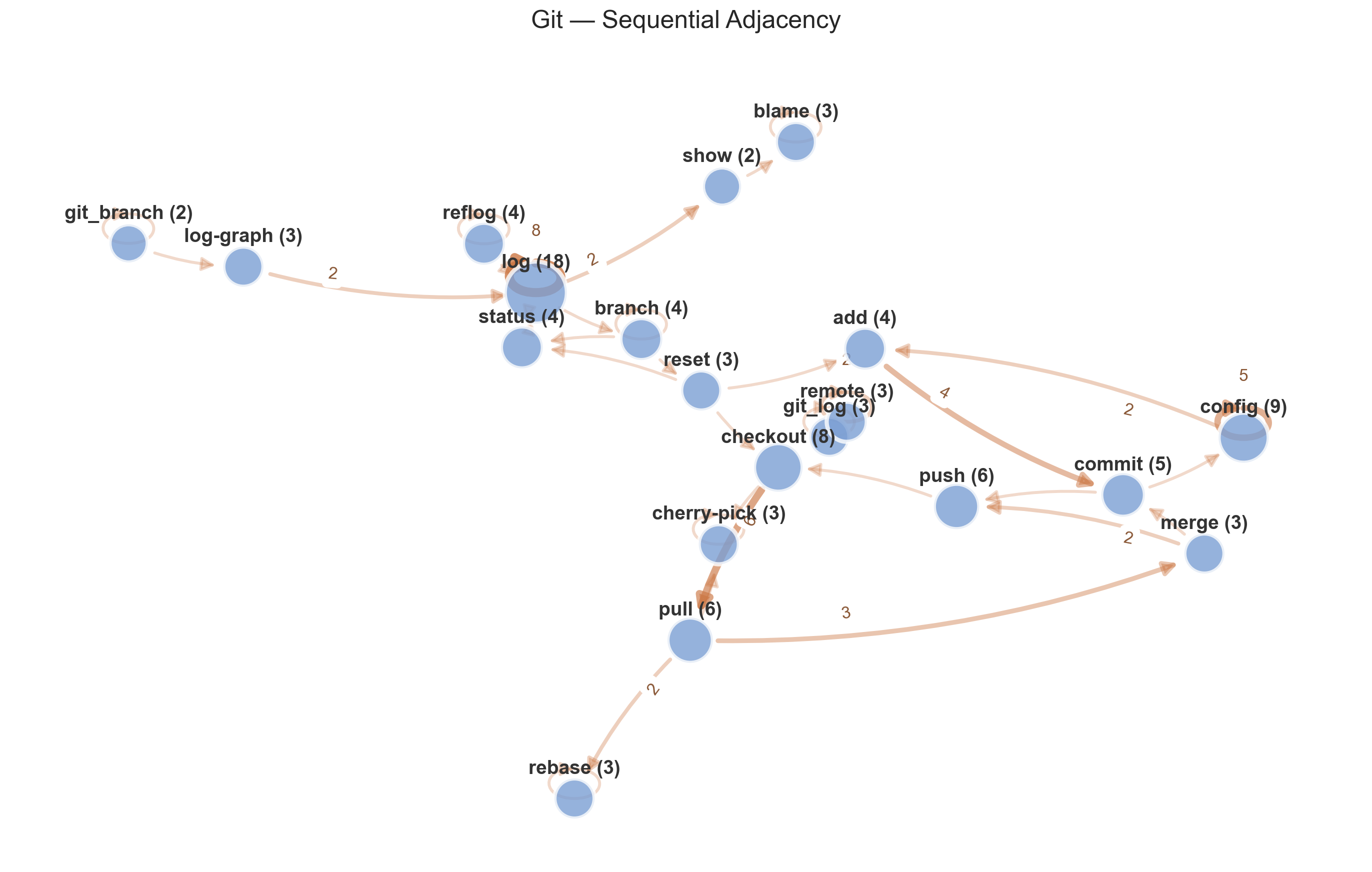}
  \caption{Sequential adjacency graph.}
\end{subfigure}
\caption{Git (33 tools): scenario categories, workflow
lengths (mean 2.2 calls), tool co-occurrence, and sequential adjacency.}
\label{fig:git}
\end{figure*}

\begin{figure*}[t]
\centering
\begin{subfigure}[t]{0.46\textwidth}
  \centering
  \includegraphics[width=\textwidth]{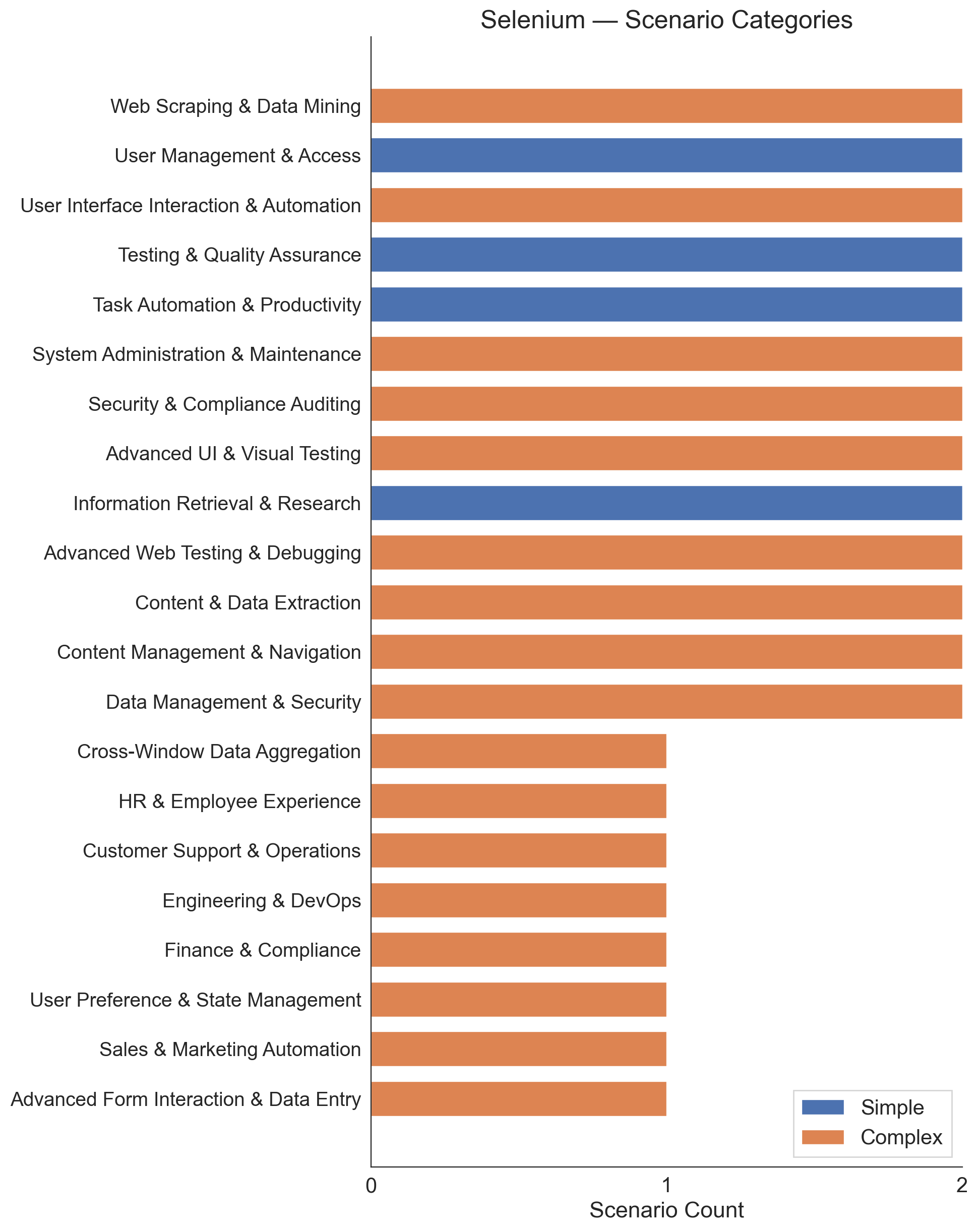}
  \caption{Scenario categories.}
\end{subfigure}
\hfill
\begin{subfigure}[t]{0.46\textwidth}
  \centering
  \includegraphics[width=\textwidth]{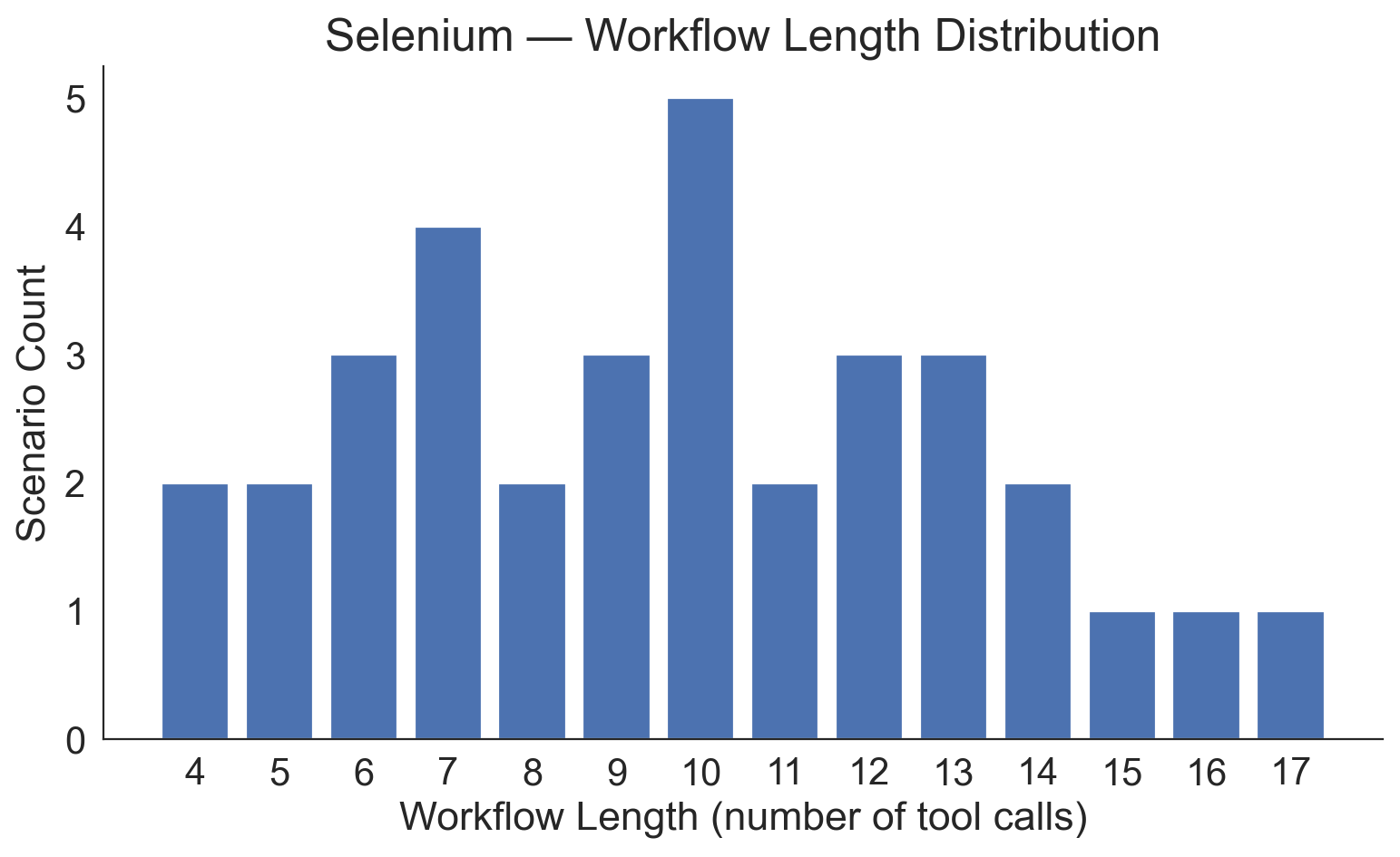}
  \caption{Workflow length distribution.}
\end{subfigure}
\\[0.5em]
\begin{subfigure}[t]{0.46\textwidth}
  \centering
  \includegraphics[width=\textwidth]{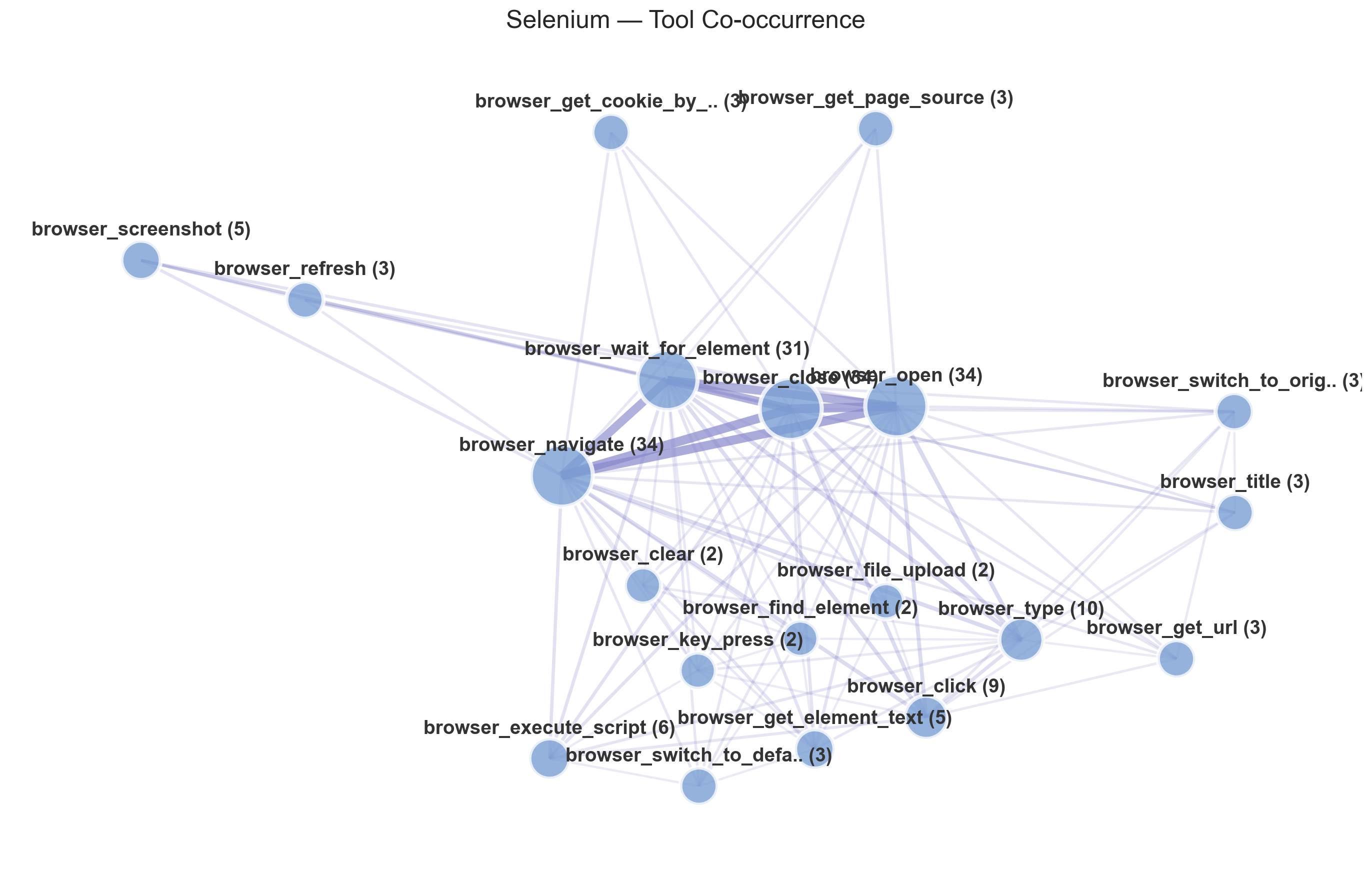}
  \caption{Tool co-occurrence graph.}
\end{subfigure}
\hfill
\begin{subfigure}[t]{0.46\textwidth}
  \centering
  \includegraphics[width=\textwidth]{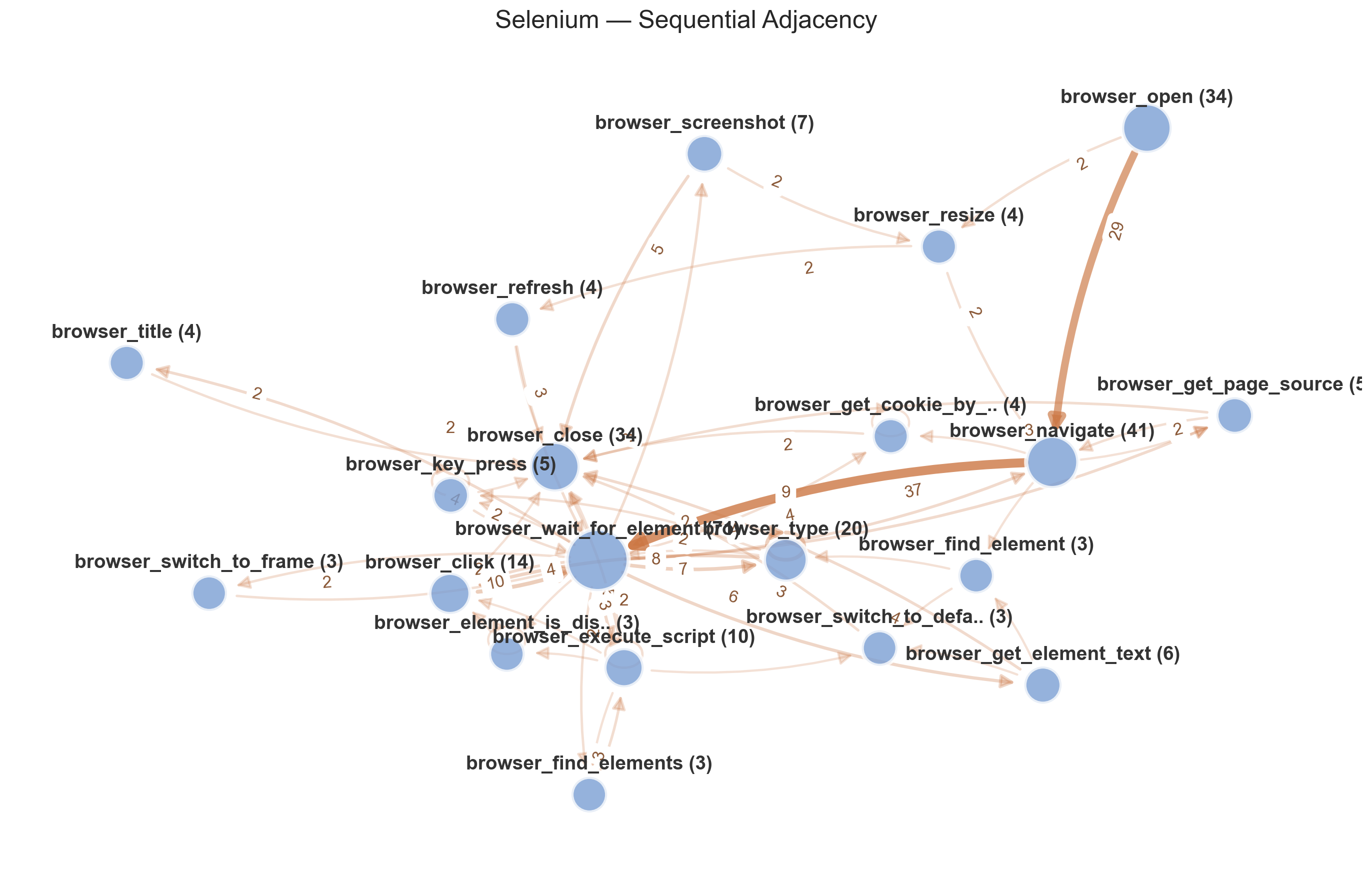}
  \caption{Sequential adjacency graph.}
\end{subfigure}
\caption{Selenium (56 tools): scenario categories, workflow
lengths (mean 9.7 calls, max 17), tool co-occurrence, and sequential adjacency.}
\label{fig:selenium}
\end{figure*}

Figures~\ref{fig:elasticsearch}--\ref{fig:selenium} show the scenario
category breakdown (simple vs.\ complex), workflow length distribution,
tool co-occurrence graph, and sequential adjacency graph for each MCP
specification. Co-occurrence graphs show which tools appear in the same
scenario (undirected); adjacency graphs show tool $A \to$ tool $B$
transition patterns within ordered workflows (directed).

\end{document}